\documentclass[letterpaper, 10pt, conference]{IEEEtran}

\IEEEoverridecommandlockouts

\ifCLASSINFOpdf
\else
\fi
\usepackage{subcaption}
\usepackage{graphicx}
\usepackage[colorlinks]{hyperref}
\usepackage{placeins}

\usepackage{tikz}
\usetikzlibrary{pgfplots.groupplots,arrows.meta,shadows,positioning,angles,quotes}
\usetikzlibrary{matrix,chains,positioning,decorations.pathreplacing,arrows}
\usetikzlibrary{shapes,arrows}
\usetikzlibrary{arrows,calc,positioning}
\usepackage{xcolor}
\usetikzlibrary{shapes.geometric}
\usetikzlibrary{positioning}
\usepackage{pgfplots}

\usepackage{forest}

\usepackage{amsmath}
\usepackage{amssymb}

\usepackage{tabularray}
\UseTblrLibrary{booktabs}

\usepackage{pifont}%
\newcommand{\cmark}{\ding{51}}%
\newcommand{\xmark}{\ding{55}}%

\newenvironment{customlegend}[1][]{%
    \begingroup
    \csname pgfplots@init@cleared@structures\endcsname
    \pgfplotsset{#1}%
}{%
    \csname pgfplots@createlegend\endcsname
    \endgroup
}%
\def\addlegendimage{\csname pgfplots@addlegendimage\endcsname}

\newcommand{\sampimg}[1]{
    \includegraphics[width = 1cm]{#1}
}

\newcommand{\sampimgtab}[1]{
    \includegraphics[height = 1.1cm]{#1}
}

\newcommand{\sampimgw}[1]{
	\includegraphics[width = \linewidth]{#1}
}

\begin{document}
\title{The Marine Debris Forward-Looking Sonar Datasets}
\author{Matias Valdenegro-Toro$^{1, 5}$ \and Deepan Chakravarthi Padmanabhan$^{2}$ \and Deepak~Singh$^{3}$ \and Bilal Wehbe$^{5}$ \and Yvan Petillot$^{4}$%
	\thanks{$^{1}$Department of Artificial Intelligence, University of Groningen, 9747AG Groningen, Netherlands.        
		{\tt\small m.a.valdenegro.toro@rug.nl}}%
	\thanks{$^{2}$Bonn-Rhein-Sieg University of Applied Sciences}
	\thanks{$^{3}$Worcester Polytechnic Institute}
	\thanks{$^{4}$Ocean Systems Lab, Heriot-Watt University}
	\thanks{$^{5}$German Research Center for Artificial Intelligence, 28359 Bremen, Germany. {\tt\small bilal.wehbe@dfki.de}}
	\thanks{This work has been partially supported by the FP7-PEOPLE-2013-ITN project ROBOCADEMY (Ref 608096) funded by the European Commission.}
}
\maketitle

\begin{abstract}
Sonar sensing is fundamental for underwater robotics, but limited by capabilities of AI systems, which need large training datasets. Public data in sonar modalities is lacking. This paper presents the Marine Debris Forward-Looking Sonar datasets, with three different settings (watertank, turntable, flooded quarry) increasing dataset diversity and multiple computer vision tasks: object classification, object detection, semantic segmentation, patch matching, and unsupervised learning. We provide full dataset description, basic analysis and initial results for some tasks. We expect the research community will benefit from this dataset, which is publicly available at \url{https://doi.org/10.5281/zenodo.15101686}.
\end{abstract}

\IEEEpeerreviewmaketitle

\section{Introduction}
Machine learning has provoked many advances in the interpretation and processing of sonar images, but these techniques are driven by the availability of large datasets, which are not always publicly available.

The machine learning and computer vision communities have many benchmark datasets that can be used for rapid prototyping and testing of new algorithms and methods, like ImageNet \cite{russakovsky2015imagenet} and OpenImages \cite{kuznetsova2020open}. Research in underwater perception and sonar image processing has similar requirements \cite{wehbe2022sonar}, but datasets are usually not publicly available due to military use or other issues. This hinders improvements in the field due to lack of standard benchmarks, and increases the barrier to newcomers into the field.

Transfer learning has been a standard approach to improve performance in computer vision systems \cite{sharif2014cnn} for a long time, but these improvements are problematic in the acoustic imaging domain, mostly due to the lack of pre-trained models trained specifically for sonar images, and more importantly, large datasets for unsupervised and self-supervised learning. Fuchs et al. \cite{fuchs2018object} uses off the shelf models pre-trained on color images (ImageNet specifically), showing that these also improve performance in sonar image classification, but these improvements are only possible due to using pre-training on ImageNet (with $>$1M images).

Most research in sonar image classification is about detecting objects of military interests, such as marine mines or wreckage, but there are many possible civilian applications. In particular we focus on human made debris sitting on the seafloor, such as plastic, metal cans, bottles, marine nets, etc.

In this paper we propose a new publicly available forward-looking sonar dataset, consisting of marine debris objects in multiple settings, and multiple tasks. The Marine Debris FLS dataset covers image classification, object detection, semantic segmentation, patch matching, and unsupervised feature learning, among other possible use cases and applications.

The dataset was captured in a laboratory (Watertank and Turntable) and real-world (Flooded Quarry) settings using a ARIS Explorer 3000 Forward-Looking Sonar, and can be used as basis for future research in understanding sonar images and development of artificial intelligence models. This paper describes the dataset in detail, including design decisions, data capture, labeling, and some initial benchmarks.

\begin{figure}
	\begin{subfigure}{0.6\linewidth}
		\includegraphics[width=\linewidth]{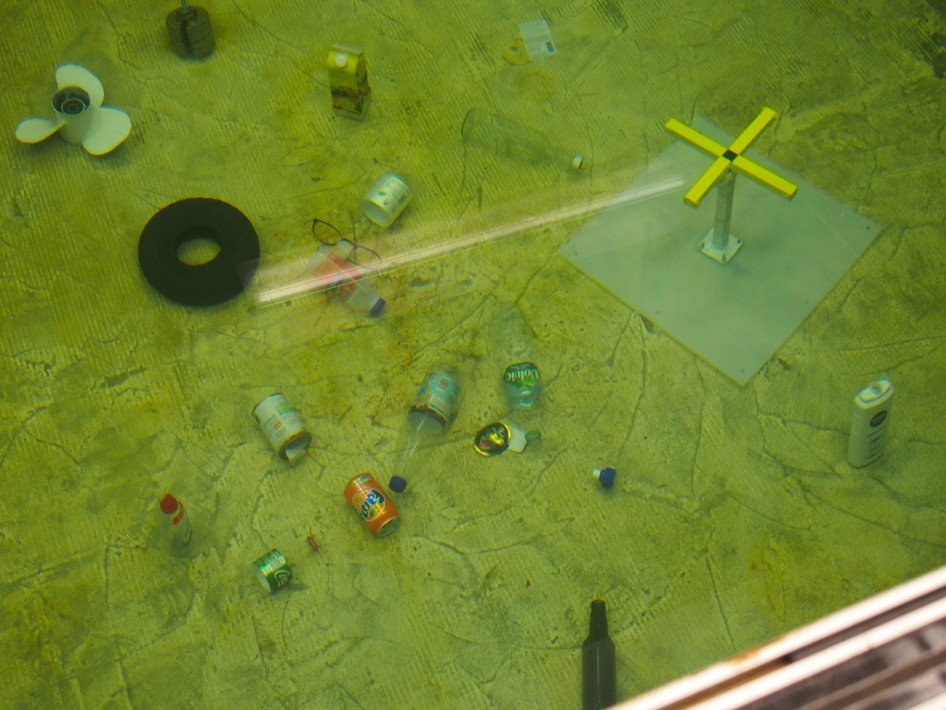}
		\caption{Scene setup.}
	\end{subfigure}
	\begin{subfigure}{0.38\linewidth}
		\includegraphics[width=\linewidth]{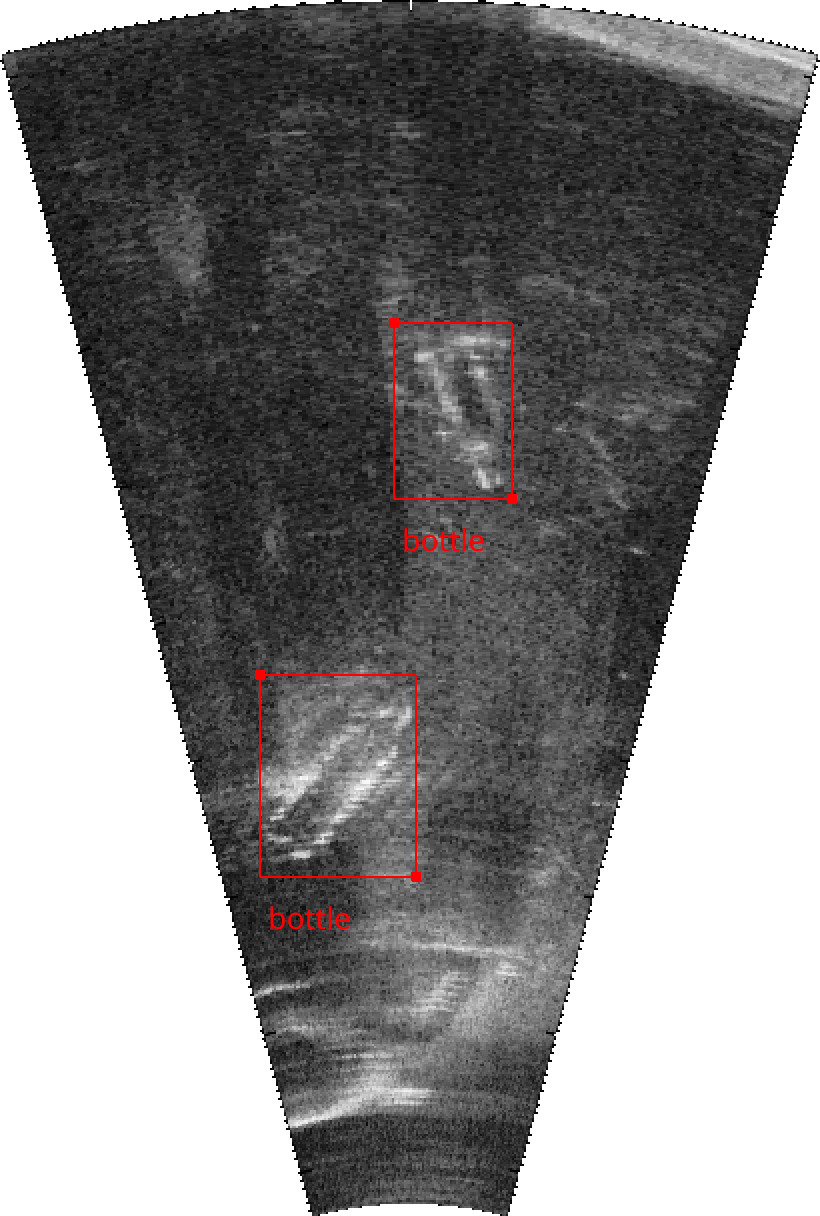}
		\caption{FLS image.}
	\end{subfigure}
	\caption{Example scene and sonar image in the watertank dataset.}
	\label{intro_fig}
\end{figure}

Contributions of this paper are:
\begin{itemize}
    \item The marine debris datasets in three different settings and up to four different computer vision tasks.
    \item One of the first publicly available sonar image datasets.
    \item Initial benchmark results showing the usefulness of our datasets.
\end{itemize}

Parts of the FLS datasets have been partially described in previous publications, like \cite{singh2021marine} for the FLS segmentation dataset, and \cite{preciado2022self} for the FLS quarry dataset, and \cite{valdenegro2021pre} for the Turntable dataset. Figures \ref{watertank_scenez}, \ref{turntable_scenez}, and \ref{quarry_scenez} visualizes the three datasets and provides basic visual information.

\begin{table*}[t]
	\begin{tblr}{Q[1.3cm, valign=b]p{6cm}p{5cm}p{4cm}}
		\toprule
							&	\textbf{Watertank} & \textbf{Turntable} & \textbf{Quarry}\\
		\midrule
		\textbf{Labels}				& Object Bounding Boxes, Semantic Segmentation,  Class Labels & Class Labels, Material Labels & None\\
		\textbf{Classes}				& Bottle, Can, Chain, Drink Carton, Hook, Propeller, Shampoo Bottle, Standing Bottle, Tire, Valve, Background (11) & Bottle, Can, Drink Carton, Box, Bidon, Pipe, Platform, Propeller, Sachet, Tire, Valve, and Wrench (12) & None \\
		\textbf{Tasks}				& Object Detection, Semantic Segmentation, Image Classification, Patch Matching & Object Classification, Material Classification & Feature Learning\\
		\textbf{Number of Samples} & 2627 & 2150 & 7209\\
		\textbf{Examples per Task} 	& \textbf{OC}: 2627, \textbf{OD}: 1868, \textbf{SS}: 1868, \textbf{PM}: 47280 & \textbf{OC}: 2150, \textbf{MC}: 2150 & \textbf{UL}: Variable\\
		\textbf{Variability}			& Low & Medium & High\\
		\textbf{Sample Images} 		& \sampimg{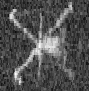} \sampimg{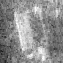} \sampimg{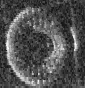}  & \sampimg{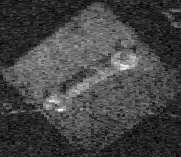}  \sampimg{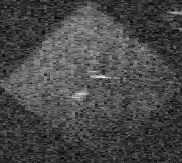} \sampimg{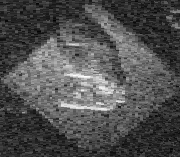} & \sampimg{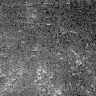} \sampimg{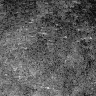} \sampimg{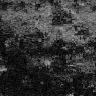}\\
		\textbf{Image  \newline Format} & Full Size and Patches & Patches & Full Size and Patches\\
		\bottomrule
	\end{tblr}
	\caption{Overview of Datasets and their related tasks: Object Classification (OC), Object Detection (OD), Patch Matching (PM), Semantic Segmentation (SS).}
	\label{dataset_overview}
\end{table*}

In this paper we publicly release and unify all datasets into a single large dataset with multiple settings and multiple tasks, as shown in Table \ref{dataset_overview}. For the Watertank dataset we have object detection and patch matching labels that are newly released, while the Turntable dataset has material recognition labels, we make full descriptions of the Turntable dataset including platform and object crops, and the Quarry dataset we release the full set of sonar frames (7209 in total) and we describe how to extract patches that are useful for unsupervised and self-supervised learning.

\section{Sonar Data Capture}

This section describes how data was captured, the three scenarios, and which object classes were used in each dataset.

\textbf{Sensor Selection}. With the objective of detecting marine debris in the seafloor, selecting an appropriate sonar sensor is crucial due to object size. Most marine debris objects are only centimetres in size (i.e. bottles, cans, drink packaging, etc), while standard sensors like sidescan sonar provide per-pixel resolutions around 10 cm per pixel.

Forward-Looking Sonar (FLS) sensors usually have higher per-pixel resolutions, in particular we selected the ARIS Explorer 3000 \cite{arisExplorer3K} as it provides resolutions up to 2.3mm per pixel (at the 3.0 MHz setting) and frame rates up to 15 FPS. Marine debris objects can clearly be seen in the output from this sensor as shown in Figure \ref{intro_fig}.

\textbf{Watertank Scenario}. As starting data, we captured a dataset the Water Tank of the Ocean Systems Laboratory, Heriot-Watt University, Edinburgh, Scotland.
The Tank measures approximately $(W, H, D) = 3 \times 2 \times 4$ [m], and on it we submerged the Nessie AUV with a sonar sensor attached to the underside, as shown in Figure \ref{intro_fig}.

Objects were placed in the water tank bottom in a pseudo-random way, without overlapping each other. Positioning was performed by dropping them into the water, and adjusting position using a pole manually.

The Nessie AUV was tele-operated to capture different views of the scene, and the water tank size limited the range of possible movements. This constraints the field of view and object poses. The vehicle performed a quarter circle sweep from side to side in order to capture all objects in the tank floor, in multiple starting positions inside the water tank.

We selected a set of household and typical objects found in marine environments, motivated that most debris objects are composed of discarded man-made objects. We also included a set of set of marine objects as distractors, which are also expected to be found in marine environments but they are not necessarily marine debris. We describe these sets below:

\textit{Household objects}. Bottles of different materials (glass, metal, plastic), tin cans, drink cartons (milk and juice), plastic shampoo bottles, a metal box, a plastic bidon, a drink sachet, and a metal wrench.

\textit{Distractors}. Chain, hook, propeller, rubber tire (two sizes), mock-up valve, plastic pipe, rotating platform.

One additional class that is always present is background, sampled from the environment outside the labeled bounding boxes for each object class. This dataset has been initially used in \cite{valdenegro2019deep}. Figure \ref{big_fls_color_objects} showcases the actual objects and their FLS patch views.

\begin{figure*}[t]
    \centering
    \begin{minipage}{0.49\linewidth}
    	\begin{subfigure}{0.49\linewidth}
    		\includegraphics[width=\linewidth]{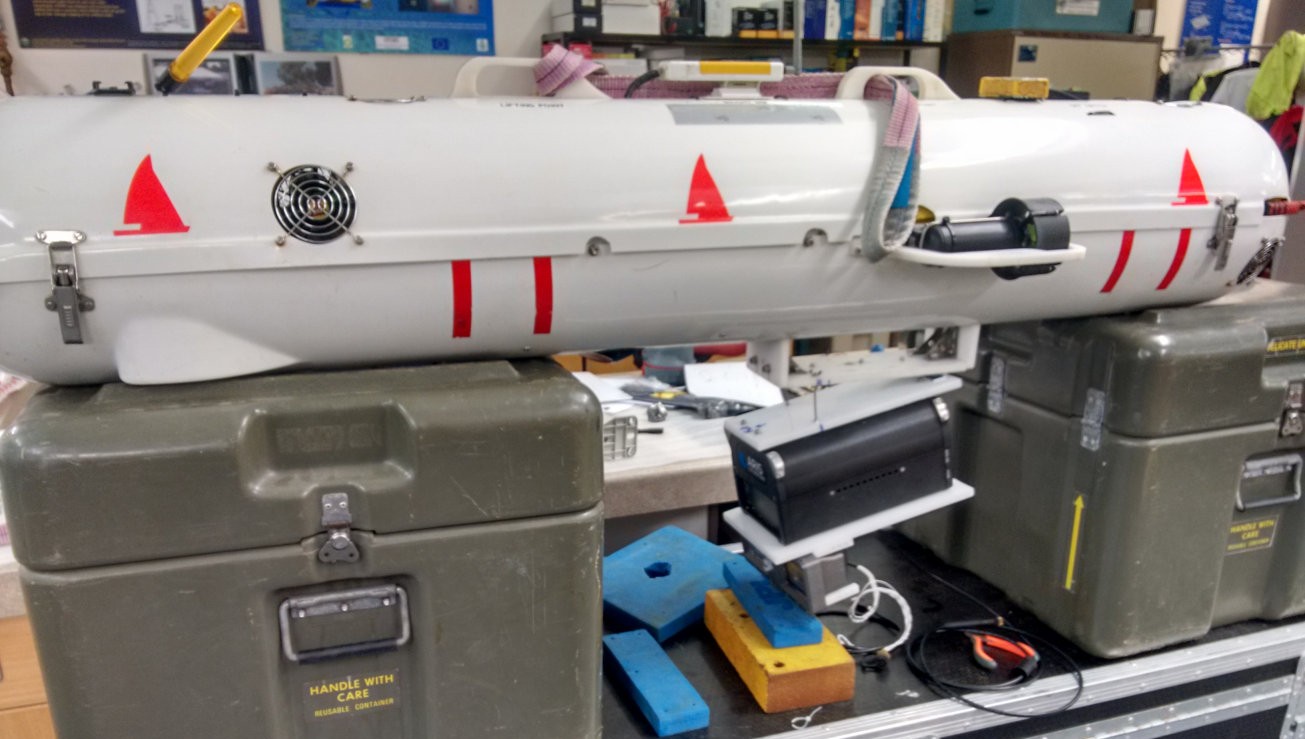}
    		\caption{The Nessie AUV with the attached ARIS Explorer 3000}
    	\end{subfigure}
    	\begin{subfigure}{0.49\linewidth}
    		\includegraphics[width = \linewidth]{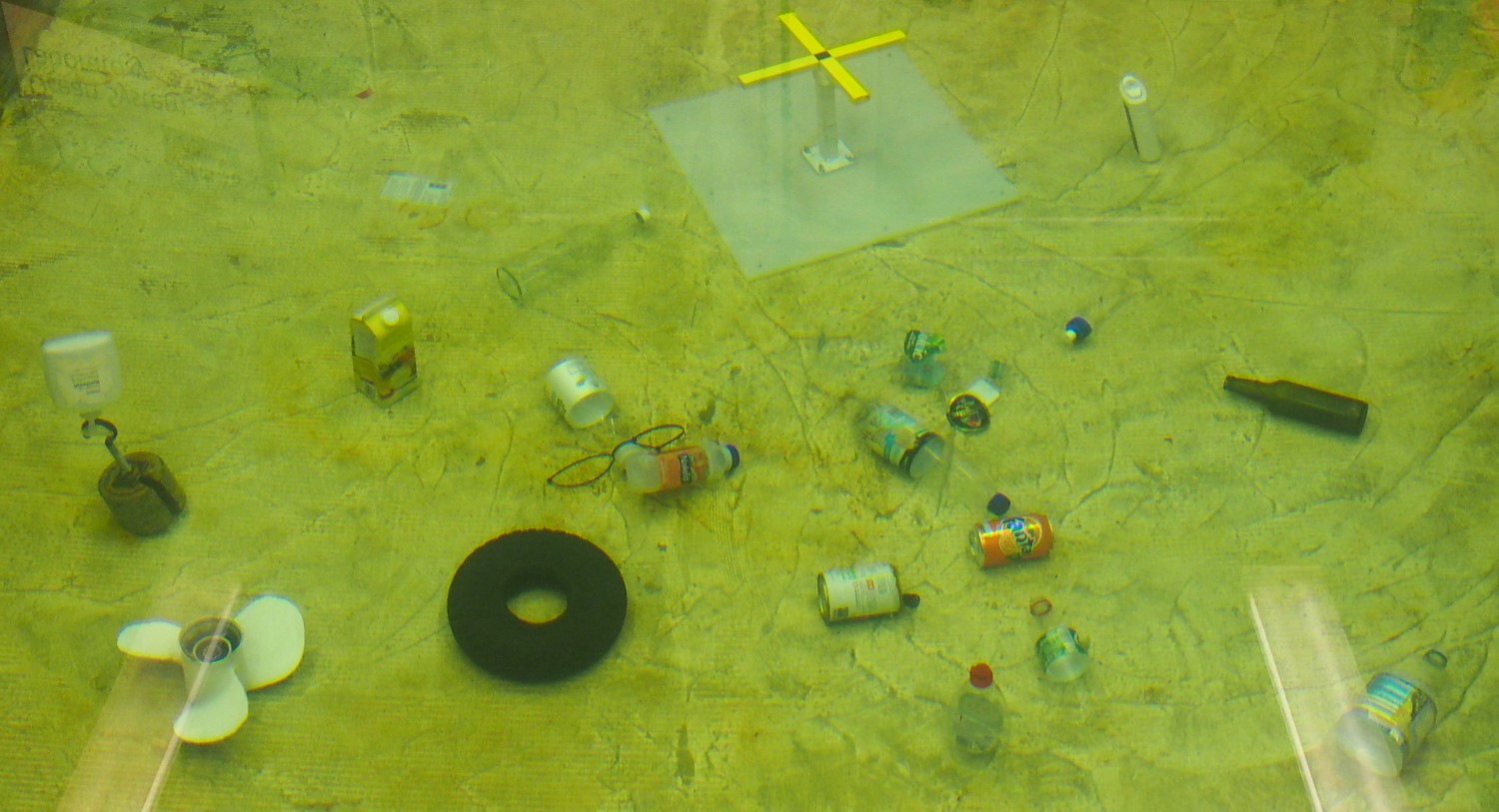}
    		\caption{View of the watertank scene with most objects in view}
    	\end{subfigure}
    	\begin{subfigure}{\linewidth}
    		\centering
    		\includegraphics[width = 0.49\linewidth]{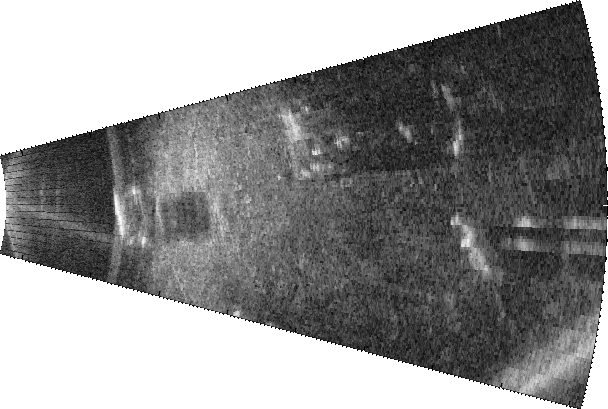}
    		\includegraphics[width = 0.49\linewidth]{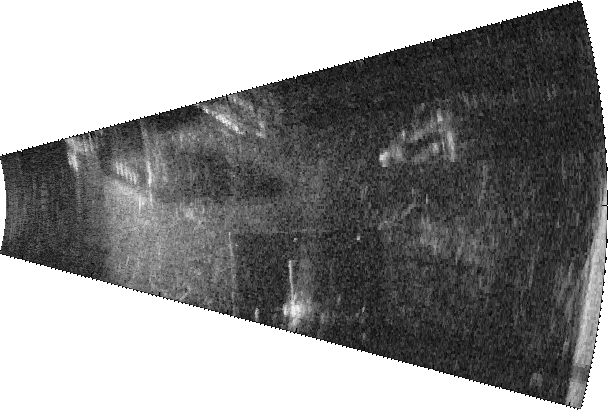}
    		\caption{Sample FLS Images}
    	\end{subfigure}
    \end{minipage}
    \begin{minipage}{0.49\linewidth}
    	\begin{subfigure}{\linewidth}
    		\includegraphics[width=\linewidth]{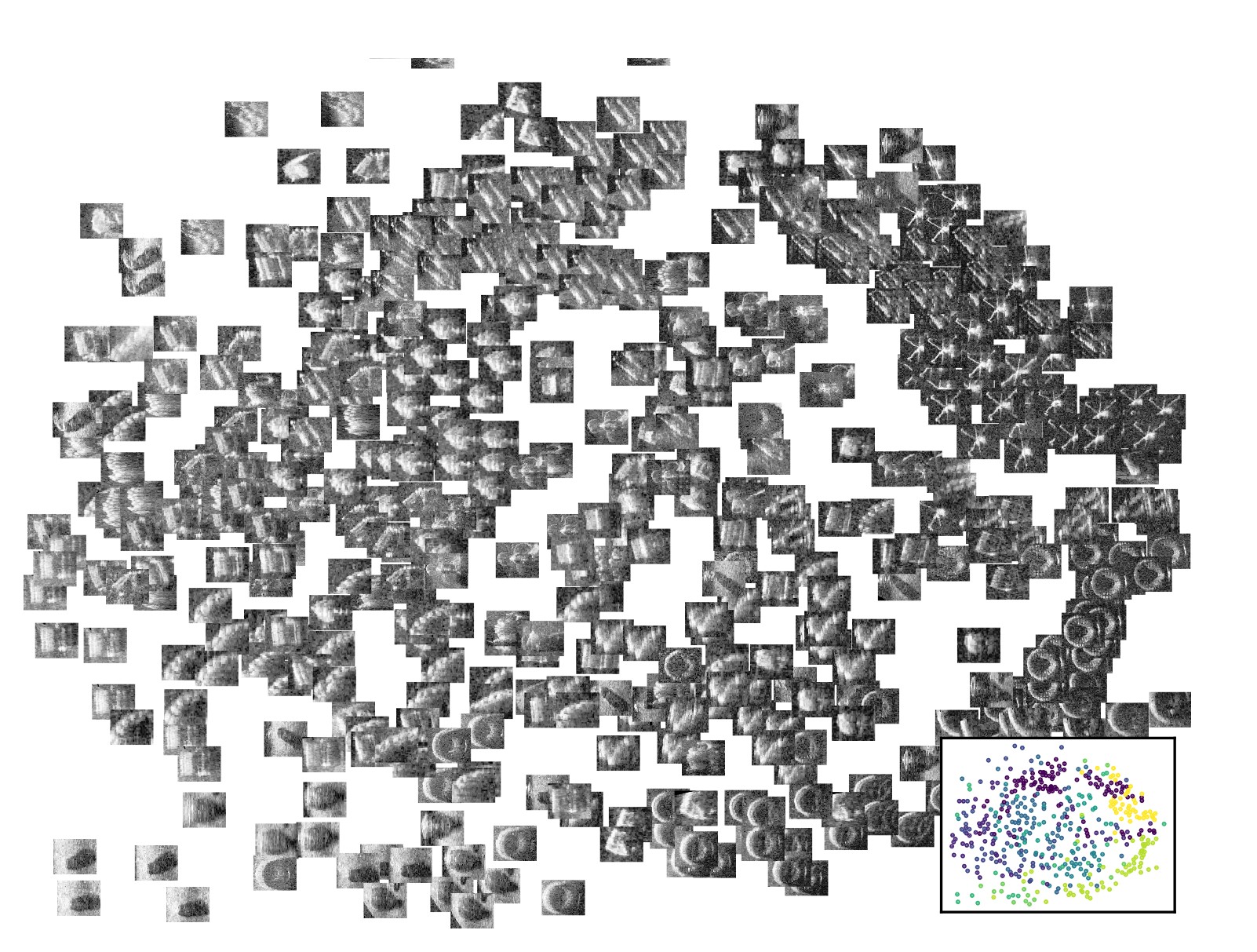}
    		\caption{MDS visualization in 2D of raw pixels, including class mapping in the bottom right.}
    	\end{subfigure}
    \end{minipage}   	   	
    \caption{Visual description of the Watertank scenario, including the AUV setup, objects in the watertank, captures FLS images, and a Multidimensional Scaling visualization of patches, showing the different classes and their pixel-space neighbors.}
    \label{watertank_scenez}
\end{figure*}

Figure \ref{watertank_scenez} shows the overall scene setup, two fullsize FLS images, and a Multidimensional Scaling (MDS) visualization of extracted patches.
    
\textbf{Turntable Scenario}. The watertank scenario has disadvantages, mainly that the variability in the images is low. Not all object poses are present in the data. Due to this we chose to create a new scenario, using a turntable to rotate objects around the Z axis, while they are underwater, and keep the sonar sensor fixed.

\begin{figure*}
	\begin{minipage}{0.53\linewidth}
		\begin{subfigure}{\linewidth}
			\includegraphics[width=0.18\linewidth]{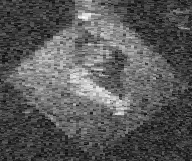}
			\includegraphics[width=0.18\linewidth]{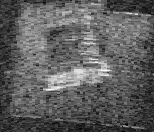}
			\includegraphics[width=0.18\linewidth]{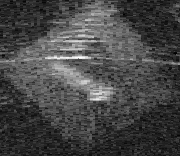}
			\includegraphics[width=0.18\linewidth]{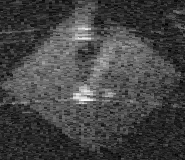}
			\includegraphics[width=0.18\linewidth]{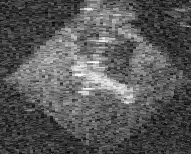}
			\caption{Different views of the metal bottle class, and how it visually changes as the turntable rotates.}
			\label{turntable_scenez_rotations}
		\end{subfigure}
		
		\begin{subfigure}{\linewidth}
			\includegraphics[width=0.32\linewidth]{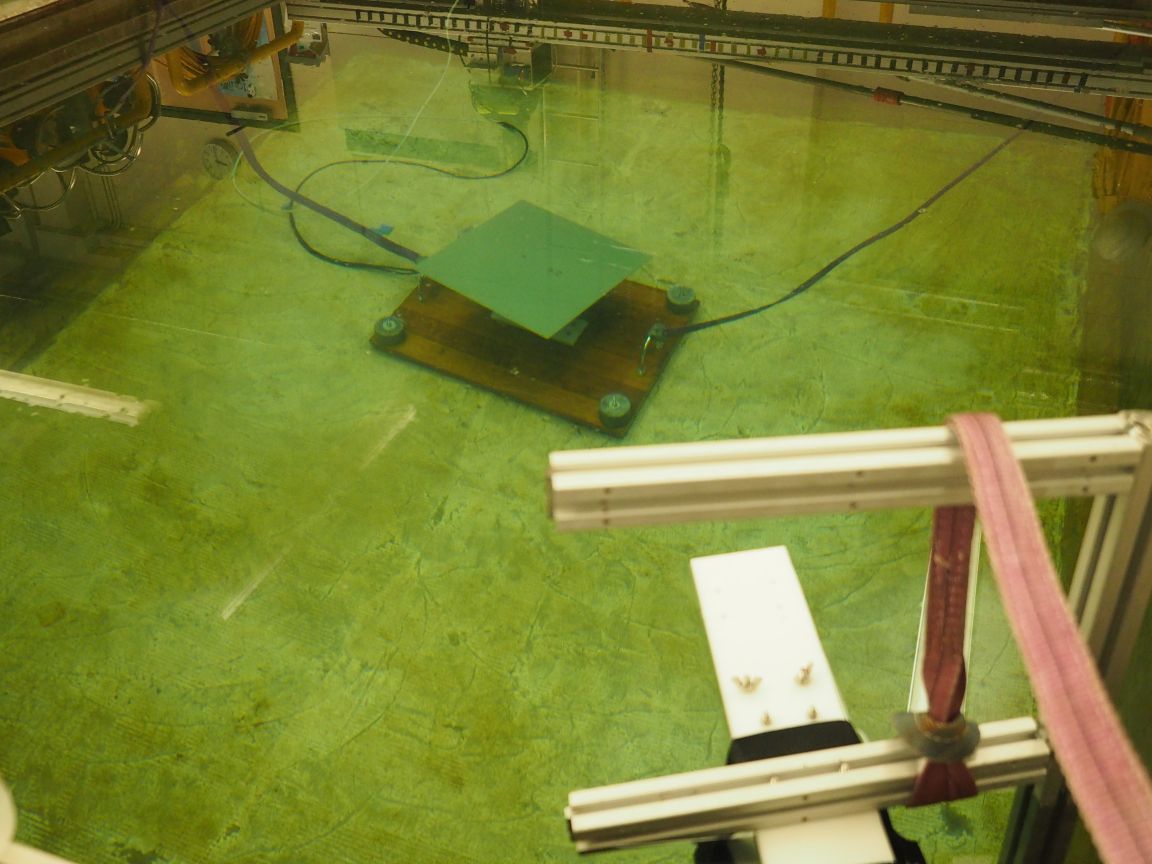}
			\includegraphics[width=0.32\linewidth]{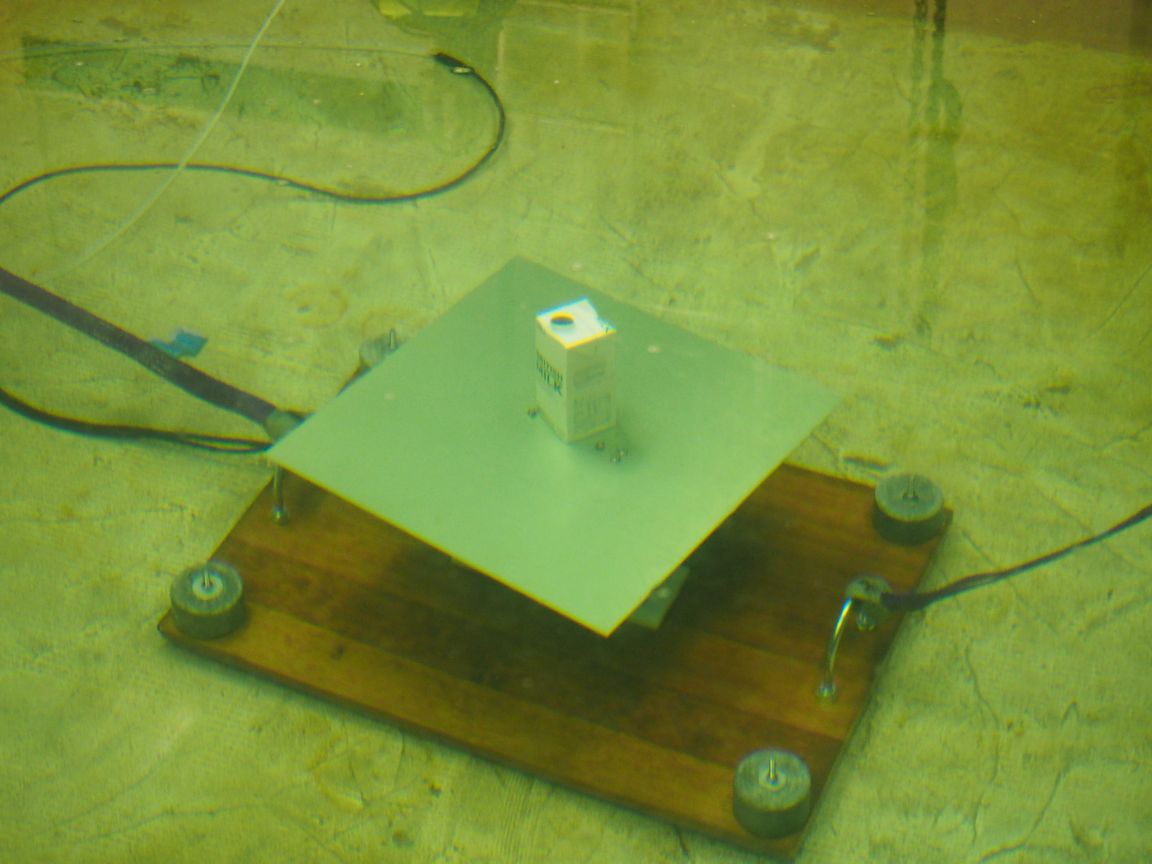}
			\includegraphics[width=0.32\linewidth]{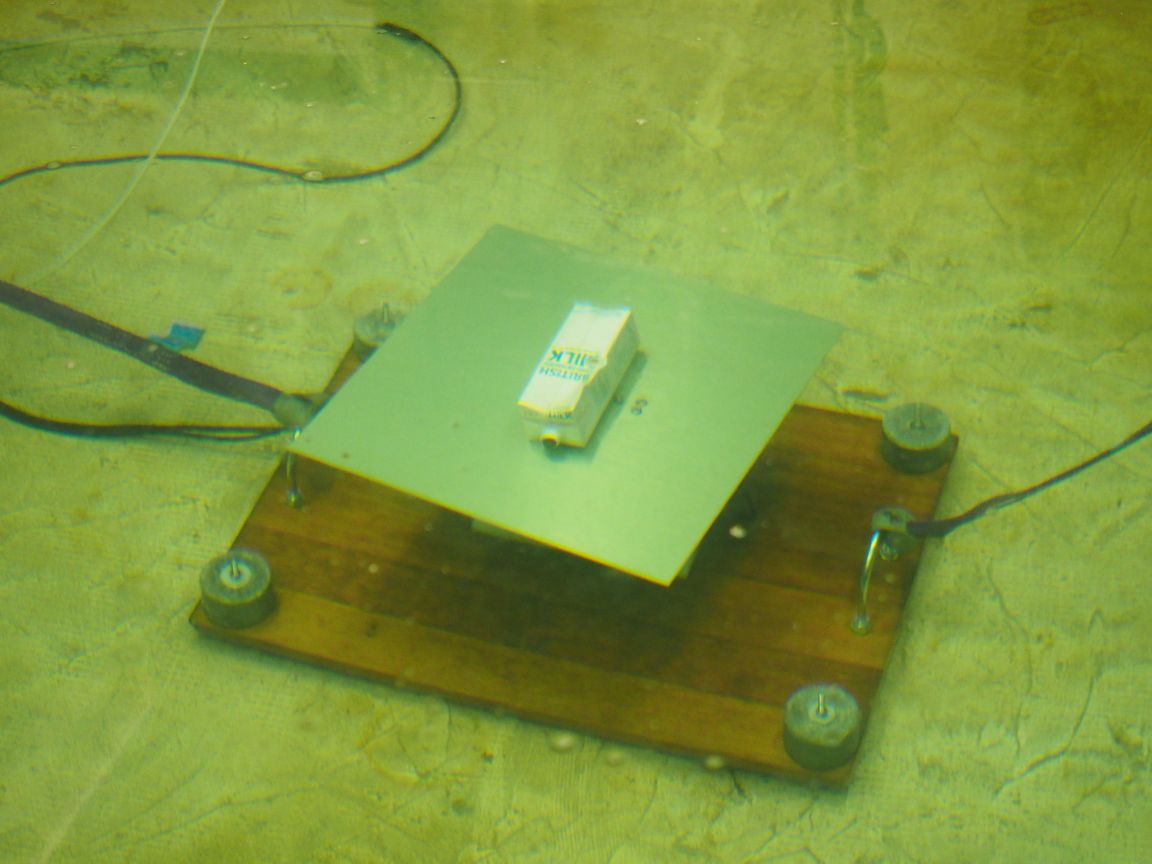}
			\caption{Turntable scene setup.}
			\label{turntable_scenez_setup}
		\end{subfigure}
	\end{minipage}
	\begin{minipage}{0.45\linewidth}
		\begin{subfigure}{\linewidth}
			\includegraphics[width=\linewidth]{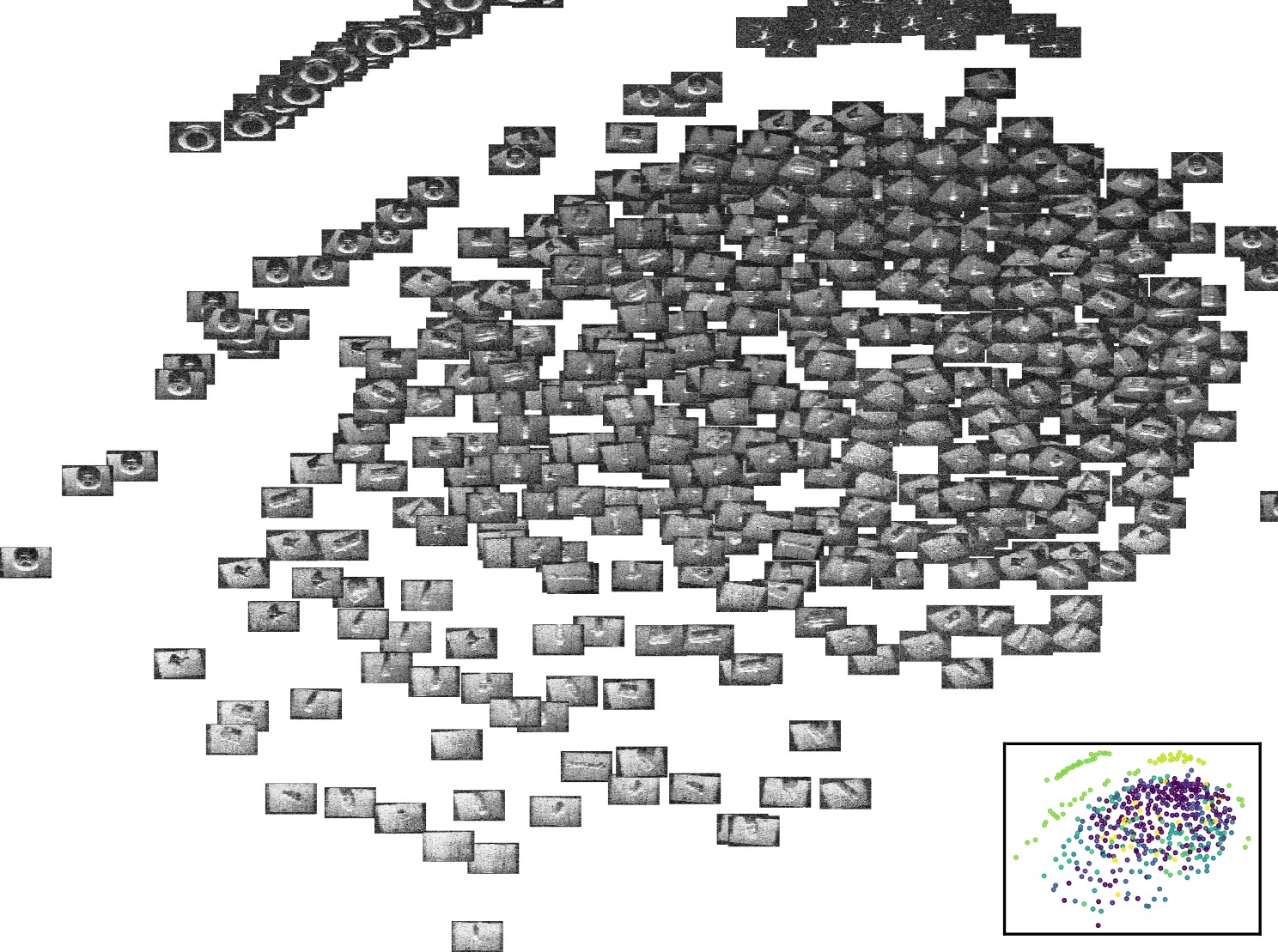}
			\caption{MDS visualization in 2D of raw pixels, including class mapping in the bottom right.}
		\end{subfigure}
	\end{minipage}		
	\caption{Visual description of the Turntable Scenario, including the Turntable setup and object placement, sample FLS images as the turntable rotates, and MDS visualization of patches, showing the increased diversity of this dataset.}	
	\label{turntable_scenez}
\end{figure*}

This setup would produce all object poses in one axis, which would increase the diversity and difficulty of this dataset. We also took the opportunity to increase the number of object classes, to: bottle, can, carton, box, bidon, pipe, platform, propeller, sachet, tire, valve, and wrench. Data generated by this scenario has been previously used in \cite{valdenegro2021pre}. The setup uses the ARIS Explorer 3000 statically mounted in a platform (shown in Figure \ref{turntable_scenez_setup}), and objects are placed in a turntable that is rotated for around $320$ degrees, producing multiple views of each object. Distance between sonar sensor and turntable is constant across all objects, around $\sim 1$mt.  Samples across rotations for one class are shown in Figure \ref{turntable_scenez_rotations}. This scenario contains both crops of the platform (which is what we use) and crops of the object inside the platform, for future use.

Some basic observations from the turntable dataset:

\textit{Artifacts}. Due to object rotation, acoustic artifacts are sometimes present, and we believe this is positive for feature learning using neural networks, unlike the watertank dataset which contains mostly static artifacts.

\textit{Shadow}. Object shadows are now clearly variable shaped, so learning algorithms are exposed to a higher variability which improves generalization and feature learning. This is shown in Figure \ref{turntable_scenez_rotations}.

\begin{figure*}[t]
	\centering
	\begin{tblr}{p{0.3cm}p{1cm}p{1cm}p{1cm}p{1cm}p{1cm}p{1cm}p{1cm}p{1cm}p{1cm}p{1cm}p{1cm}p{1cm}}
		\toprule
		& Bidon & Bottle & Box & Can & Drink Carton & Pipe & Platform & Propeller & Sachet & Tire & Valve & Wrench\\
		\midrule
		\rotatebox[origin=r]{90}{Optical} & \sampimgw{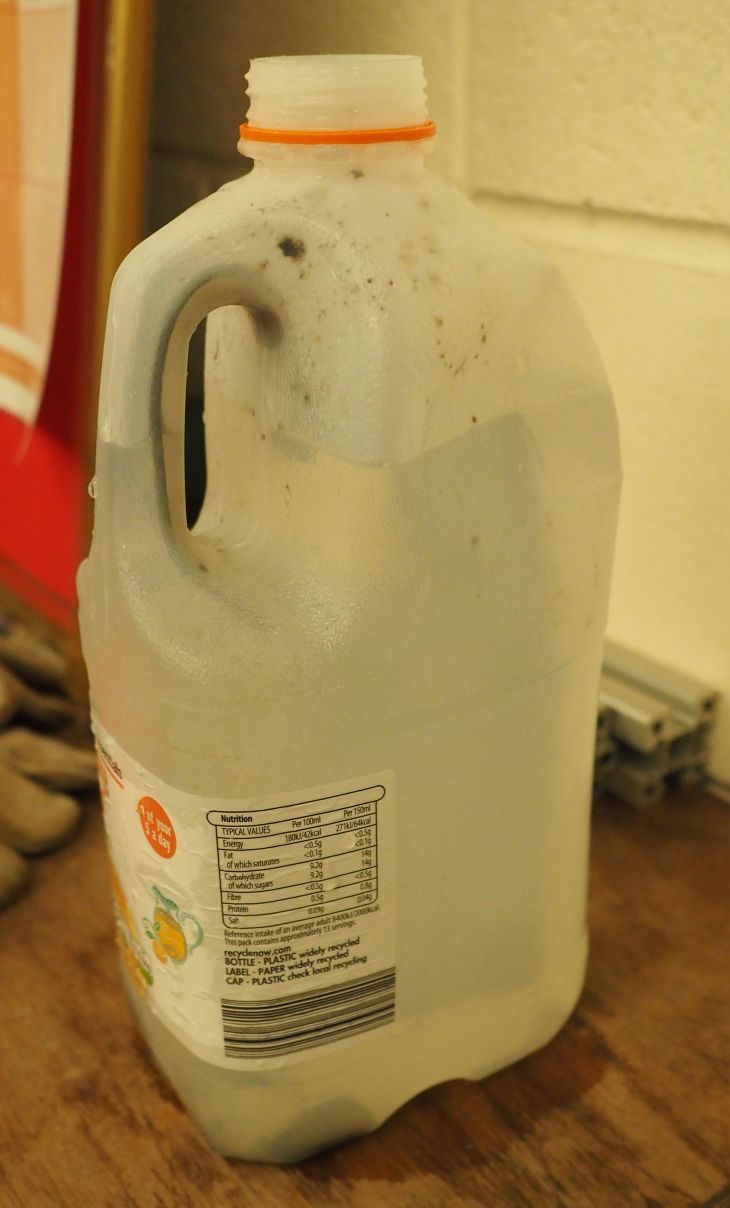} & \sampimgw{images/watertank/glass-bottle.jpg} & \sampimgw{images/turntable/box} & \sampimgw{images/watertank/can} & \sampimgw{images/watertank/drink-carton} & \sampimgw{images/turntable/pipe} & \sampimgw{images/turntable/turntable} & \sampimgw{images/watertank/propeller} & \sampimgw{images/turntable/drink-sachet} & \sampimgw{images/watertank/tire} & \sampimgw{images/watertank/valve} & \sampimgw{images/turntable/wrench}\\
		
		\rotatebox[origin=r]{90}{FLS SW} & \sampimgw{images/turntable/classes/plastic-bidon-sideways} & \sampimgw{images/turntable/classes/plastic-bottle-sideways} &  & \sampimgw{images/turntable/classes/can-sideways} & \sampimgw{images/turntable/classes/drink-carton-sideways} & \sampimgw{images/turntable/classes/plastic-pipe-sideways} & & \sampimgw{images/turntable/classes/plastic-propeller-sideways} & \sampimgw{images/turntable/classes/drink-sachet-sideways} &  & & 
		\\
		
		\rotatebox[origin=r]{90}{FLS ST} &  & \sampimgw{images/turntable/classes/plastic-bottle-standing} & \sampimgw{images/turntable/classes/metal-box-standing} & \sampimgw{images/turntable/classes/can-standing} & \sampimgw{images/turntable/classes/drink-carton-standing} & &  \sampimgw{images/turntable/classes/rotating-platform-standing} & & & \sampimgw{images/turntable/classes/small-tire-standing} & \sampimgw{images/turntable/classes/valve-standing} & \sampimgw{images/turntable/classes/wrench-standing}
		\\
		\bottomrule
	\end{tblr}
	\caption{Sample color and FLS images from the turntable dataset, showcasing the real-world object that were used to build this dataset. We include both sideways (SW) and standing (ST) views when applicable.}
	\label{turntable_objects_color_fls}
\end{figure*}

Figure \ref{turntable_scenez} shows the overall setup, some FLS images as the turntable rotates, and a MDS visualization showcasing the diversity in this dataset. Figure \ref{turntable_objects_color_fls} showcases the real objects and their FLS views.

\textbf{Flooded Quarry Scenario}. The last scenario consists of data we captured in a flooded quarry in the Edinburgh, Scotland area. This dataset is meant to be used for unsupervised learning, so there are no labels, which allows more freedom in data capture. The dataset is larger than previous ones, at 7004 full size FLS images.

In this scenario, we attached the ARIS Explorer 3000 in a floating platform, pointing down approximately $30^\circ$ from the horizontal. This platform was towed using a surface vehicle (a EvoLogics Sonobot) at a slow constant speed, mapping the quarry/lake bottom. Many images contain empty areas (denoted as full black), so from the total sonar scans that were captured, we filter black regions out to keep useful areas with lake bottom.

Figure \ref{quarry_scenez} presents premade patch datasets that we provide in this paper, but it is possible to produce patches at other sizes or strides to obtain larger datasets. We also build an MDS visualization to show patch diversity, and showcase some individual patches.

\section{Marine Debris FLS Datasets}
This section describes the datasets and tasks that derive from the three basic scenarios, from a machine learning perspective. An overview of all datasets and tasks in presented in Table \ref{dataset_overview}.

\subsection{Tasks}
The tasks available in the marine debris datasets are:

\textbf{Object Classification}. Each image contains a single instance of an object class, and the objective is for model to predict the correct class from the set of known classes. Watertank and Turntable datasets support this task, and is evaluated using categorical accuracy.

\textbf{Patch Matching}. Two images of the same dimensions ($96 \times 96$) are given as input, and the task is to determine if the two images contain the same object or not, which is binary classification. The challenge is that each image might represent objects in different views, or be completely different objects or background. This task is available in the Watertank dataset, and is evaluated using binary accuracy, area under the ROC curve (AUROC), or other binary classification metrics. For training and evaluation we provide two settings: same or different objects between train and validation, for more information please see \cite{valdenegro2017improving}.

\textbf{Material Classification}. Similar to the object classification task, but the ground truth labels are defined as material classes instead of visual object classes. This task is more difficult as material is an implicit property not directly visually observed, but acoustic returns should have information about physical materials, due to acoustical physics travelling across different materials. This task is evaluated using categorical accuracy.

\textbf{Object Detection}. Given a full sonar image in rectangular representation, containing a certain number of objects of interest (might be zero), the task is to produce a bounding box with a classification decision for each object in the image. Objects are only present inside the sonar's field of view. This task is only available in the Watertank dataset, and is evaluated with the Mean Average Precision (mAP).

\textbf{Semantic Segmentation}. Same setting as the object detection task, but instead the task is to assign a classification decision to each pixel inside the sonar's field of view. This task is only available in the Watertank dataset, and is evaluated by the per-class Intersection-over-Union (IoU), mean IoU over all classes, and Dice score.

\textbf{Unsupervised Learning}. This is an open ended task, only available in the Flooded Quarry dataset, while in theory it could be applied to Watertank and Turntable datasets, if the labels are discarded. Possible tasks are to learn useful feature representations, self-supervised learning, clustering, image generation, density estimation, etc. The advantage of unsupervised learning is only possible in large datasets, the two previous datasets are not large as the Flooded Quarry dataset. We expect that a typical example of this task is feature learning via self-supervised methods. Evaluation depends on the specific unsupervised task, for feature learning the most common evaluation is performance on a downstream task via transfer learning.

\begin{figure*}[t]
	\centering
	\begin{subfigure}{0.59\linewidth}
		\includegraphics[width=0.40\linewidth, angle=90]{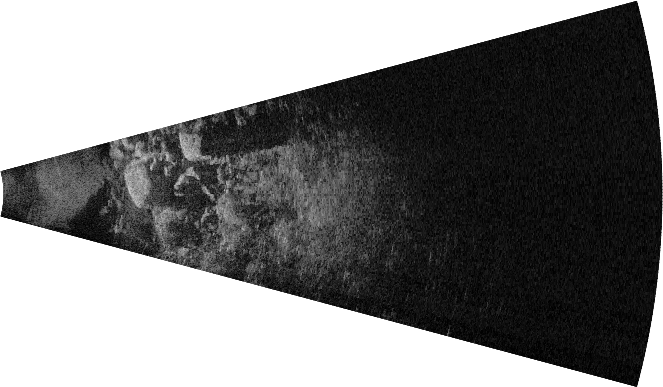}
		\includegraphics[width=0.40\linewidth, angle=90]{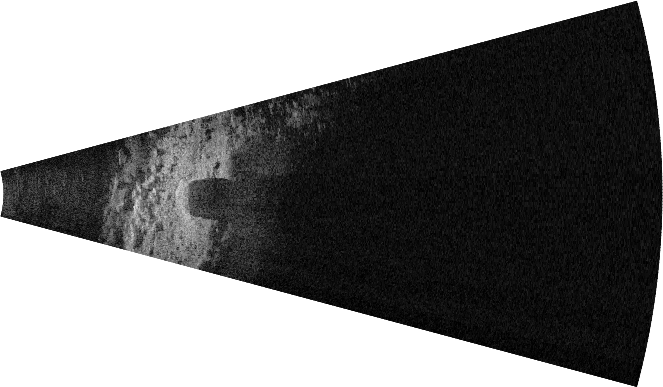}    
		\includegraphics[width=0.40\linewidth, angle=90]{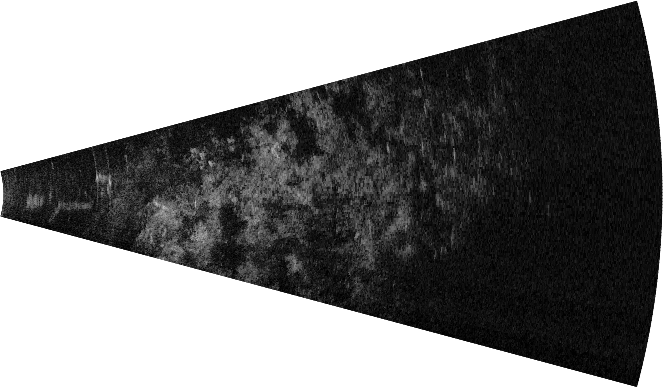}
		\includegraphics[width=0.40\linewidth, angle=90]{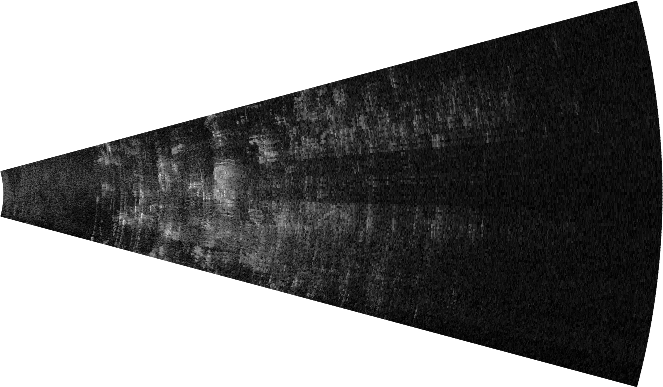}
		\caption{Sample full size FLS Images}
		\label{quarry_scenez_fls}
	\end{subfigure}
	\begin{subfigure}{0.39\linewidth}
		\includegraphics[width=\linewidth]{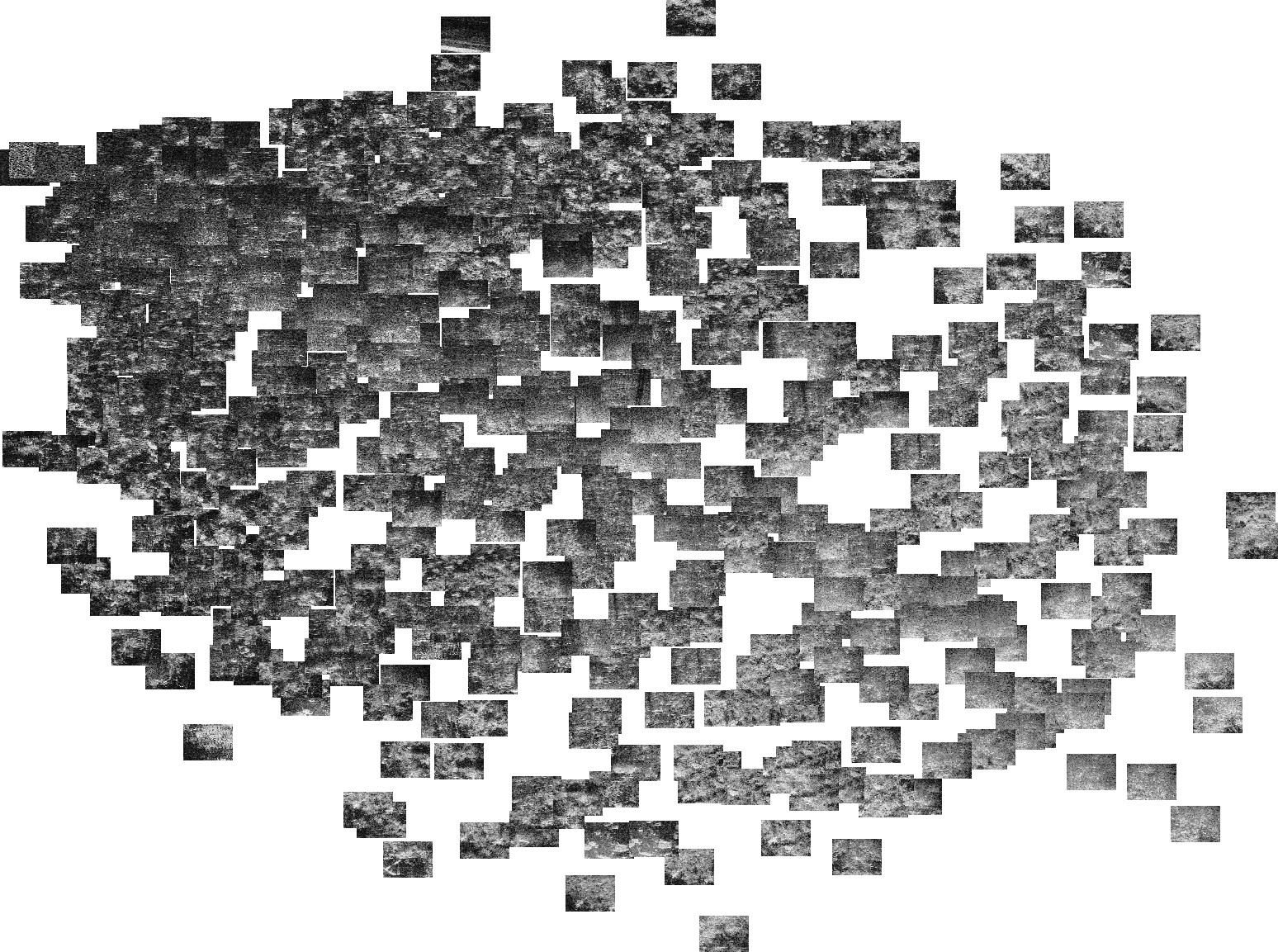}
		\caption{MDS visualization in 2D of raw pixel patches.}
	\end{subfigure}
	
	\begin{subfigure}{0.59\linewidth}
		\begin{tabular}{llllll}
			\toprule
			Patch $W \times H$		& $S$ 	& \# of Samples & Patch $W \times H$ 	& $S$ 	& \# of Samples\\
			\toprule
			$128 \times 128$	& 4			& 57639 & $96 \times 96$		& 16		& 48760\\
			$128 \times 128$	& 16		& 37352 & $64 \times 128$		& 2			& 68534\\
			$96 \times 96$		& 32		& 32185\\						
			\bottomrule
		\end{tabular}
		\caption{Unlabeled Dataset Size as function of stride and patch size}
		\label{quarry_patches_sizes}
	\end{subfigure}
	\begin{subfigure}{0.39\linewidth}
		\centering
		\includegraphics[width=0.18\linewidth]{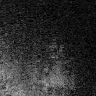}
		\includegraphics[width=0.18\linewidth]{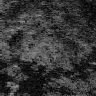}
		\includegraphics[width=0.18\linewidth]{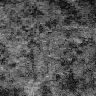}
		\includegraphics[width=0.18\linewidth]{images/quarry/samples/patch-24149.png}
		\vspace*{0.1em}
		
		\includegraphics[width=0.18\linewidth]{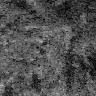}
		\includegraphics[width=0.18\linewidth]{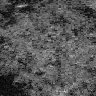}
		\includegraphics[width=0.18\linewidth]{images/quarry/samples/patch-31590.png}
		\includegraphics[width=0.18\linewidth]{images/quarry/samples/patch-6335.png}
		\caption{Sample patches at $W, H = 96 \times 96$}
	\end{subfigure}
	\caption{Visual description of the Quarry scenario, including fullsize FLS images, premade patch dataset statistics, MDS visualization, and some sample patches.}
	\label{quarry_scenez}
\end{figure*}

\subsection{Labeling}

\begin{figure*}
	\begin{forest}
		[Marine Debris
		[Plastic
		[Bottle]
		[Drink Sachet]
		[Pipe]
		]
		[Metal
		[Box]
		[Can]
		[Chain]
		[Hook]
		[Wrench]
		]
		[Composite
		[Platform]
		[Valve]
		]            
		[Rubber
		[Tire]
		]
		[Glass
		[Bottle]
		[Jar]
		]
		]        
	\end{forest}
	\caption{Hierarchy of Labels. First level is material, while the leafs are object classes.}
	\label{label_hierarchy}
\end{figure*}
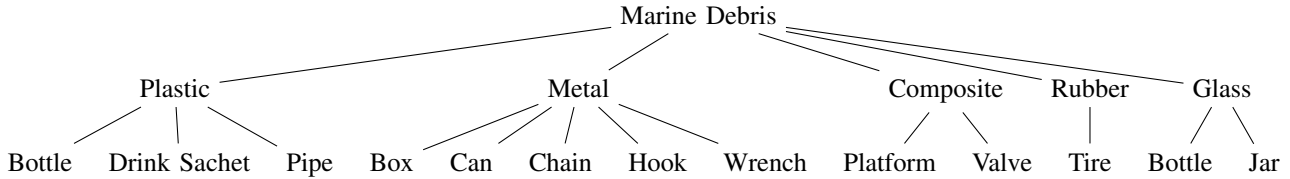

\textbf{Objects}. We provide object and material annotations. We first labeled object classes manually by visual inspection, as we have a controller environment (Watertank and Turntable) where objects are known, and then from object classes, we can infer the material labels as shown in the label hierarchy in Figure \ref{label_hierarchy}. There is one object that is available in two materials (glass and plastic bottle), but they can be differentiated by the object class name which includes the material.

\textbf{Match Decisions}. These are binary labels indicating match or no match, and they are automatically generated by sampling combinations of patches between object classes and background. From the Watertank dataset, from FLS full size images, we sample image patches, including background. Then patches are randomly grouped into pairs, and if two patches have the same object class, this is labeled as a match (label equal to 1.0), and if they have different object classes or one of them is background, this is labeled as a non-match (label equal to 0.0).

\textbf{Bounding Boxes}. These are per-object bounding boxes with object information for object detection. These were manually labeled by an expert, tightly enclosing each recognizable object infull-size sonar images inside the Watertank dataset. These labels are available as XML files in PASCAL VOC format.

\textbf{Segmentation Masks}. They are per-pixel object class annotations for semantic segmentation, they were labeled by an expert using polygons over the image which are rasterized to produce a mask image, encoding classes as integer values. More details are available in \cite{singh2021marine}.

\subsection{Preprocessing}

Most datasets presented in this paper require some kind of preprocessing before being usable for a machine learning or computer vision task. In this section we provide the best practices we recommend.

\textbf{Watertank and Turntable Scenarios}. Tasks in these scenarios do not require complicated preprocessing, usually performing min-max normalization, or just dividing the image pixels by 255, is enough to successfully train models.

\textbf{Quarry Scenario}. It is possible to use this dataset in rectangular format (as shown in Figure \ref{quarry_scenez_fls}), but more often for feature learning, patches must be extracted from these images. We propose a standard formulation to extract patches. We generate a set of windows of size $(W, H)$ with a stride\footnote{Pixel step between windows} $S$ that are only present in the sonar's polar field of view, and then each window extracts one patch from the full size sonar image. Varying values of $W, H, S$ will generate different number of patches. To reduce the number of empty patches due to intersection between the acoustic beam and the environment, we remove any window where any of the four corners or the center point have no acoustic return, defined as an image value below 20. We provide prebuilt patch datasets using this method at varying patch sizes and strides in our github repository, a summary is presented in Table \ref{quarry_patches_sizes}.

\textbf{Full Size Rectangular FLS Images}. Watertank and Quarry scenarios are preprocessed to be usable as image patches for most tasks, but the full size rectangular FLS images are also available (as shown in Figure \ref{quarry_scenez_fls}) for future use, for example for unsupervised learning or other applications. These images contain a rectangular projection of the FLS' polar field of view, which vary between images. Empty pixels (black or zero value) in the area surrounding the polar field of view are as padding/empty areas. To only process pixels inside the field of view, a flood fill algorithm can be used to produce a mask for pixels surrounding the polar field of view and filter these away.

\textbf{Patch Size}. Datasets are available at multiple patch sizes (usually $32 \time 32, 64 \time 64, 96 \time 96, 128 \times 128$), while full size sonar images have variable sizes. For machine learning models, the best is a constant image size, and experimentally we have found the best trade-off for many tasks is $96 \ times 96$ pixels.

\subsection{File Structure and Splits}

\textbf{Image Patches}. Object classification, Patch Matching and Unsupervised Patch Learning datasets are individual image patches, and are stored as HDF5 files in grayscale format. When applicable, labels are stored as integers and class names are available as an attribute.

\textbf{Full Size Rectangular FLS Images}. Object Detection, Semantic Segmentation, and Unsupervised Learning (outside of patches) require full size FLS images, these are available as PNG files. Some preprocessing is required to focus algorithms into the usable polar field of view, as previously described.

\subsection{Licensing}. All data presented in this paper is available under a Creative Commons Non-Commercial Share Alike license, for other uses and specific licensing, please contact Heriot-Watt University and the Robocademy project consortium, in particular the German Research Center for AI in Bremen, Germany.

\begin{figure*}[t]
	\centering
	\begin{tblr}{Q[1cm, valign=t]p{1cm}p{1cm}p{1cm}p{1cm}p{1cm}p{1cm}p{1cm}p{1cm}p{1cm}p{1cm}}
		\toprule
		& Bottle & Can & Chain & Drink Carton & Hook & Propeller & Shampoo Bottle & Standing Bottle & Tire & Valve\\
		\midrule
		Color Image & \sampimgw{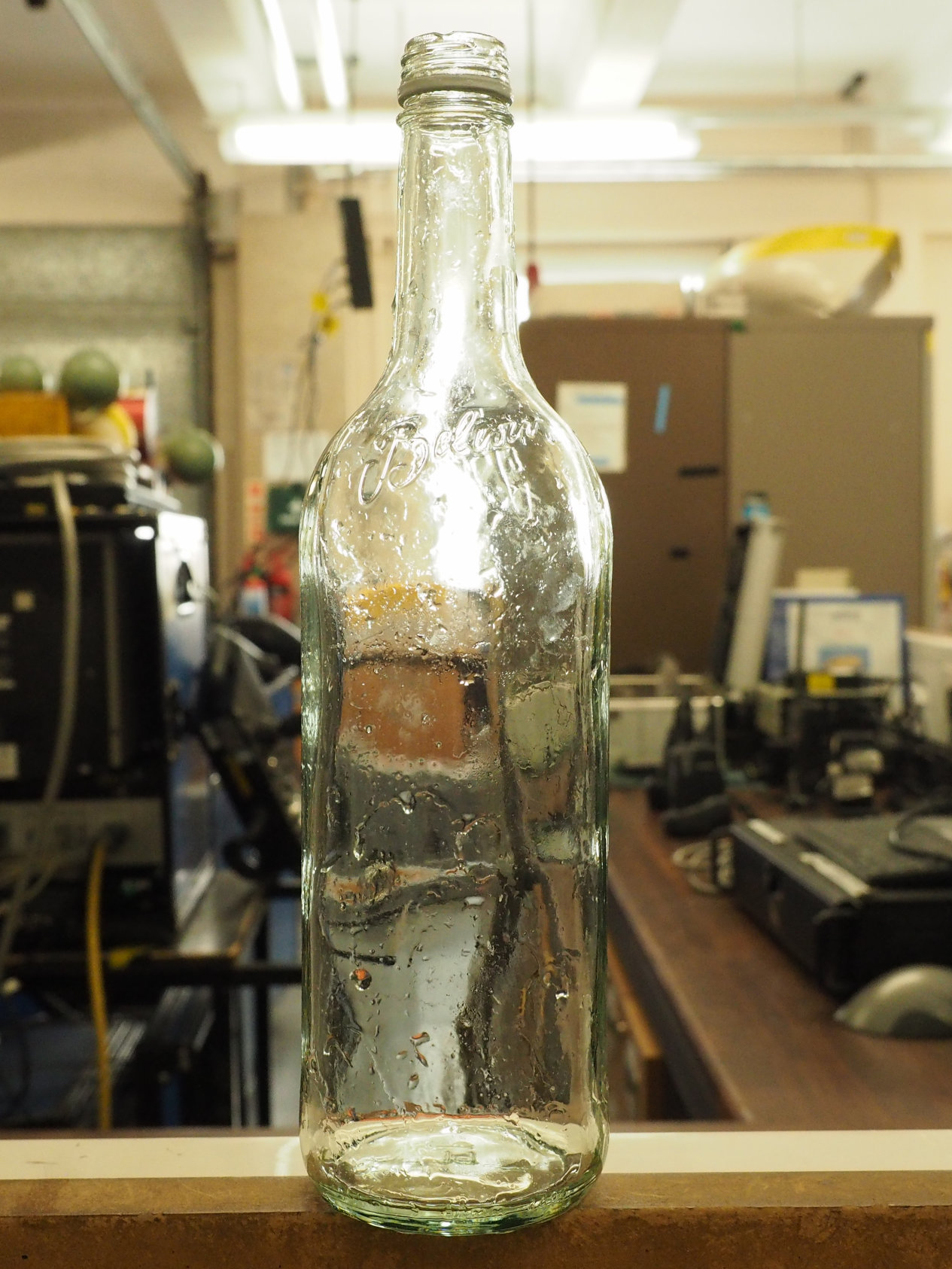} & \sampimgw{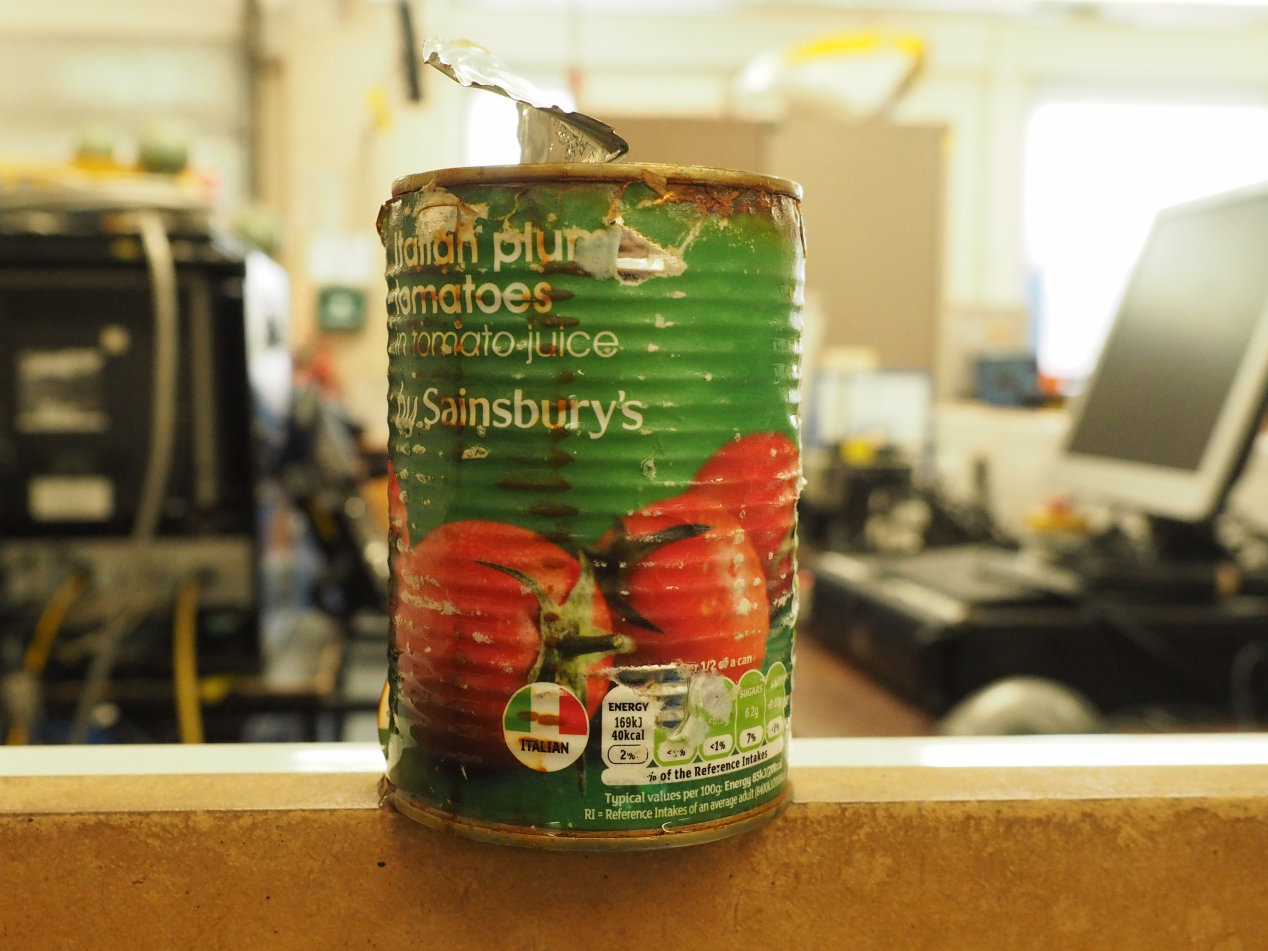} & \sampimgw{images/watertank/chain} & \sampimgw{images/watertank/drink-carton} & \sampimgw{images/watertank/hook} & \sampimgw{images/watertank/propeller} & \sampimgw{images/watertank/shampoo-bottle} & \sampimgw{images/watertank/beer-bottle} & \sampimgw{images/watertank/tire} & \sampimgw{images/watertank/valve}\\
		FLS Image & \sampimgw{images/watertank/classes/bottle-0} & \sampimgw{images/watertank/classes/can-0} & \sampimgw{images/watertank/classes/chain-0}& \sampimgw{images/watertank/classes/drinkCarton-0} & \sampimgw{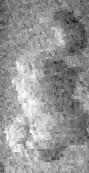} & \sampimgw{images/watertank/classes/propeler-0} & \sampimgw{images/watertank/classes/shampooBottle-0} & \sampimgw{images/watertank/classes/standingBottle-0} & \sampimgw{images/watertank/classes/tire-0} & \sampimgw{images/watertank/classes/valve-0}\\
		\bottomrule
	\end{tblr}
	\caption{Sample color and FLS images from the watertank dataset, showcasing the real-world object that were used to build this dataset.}
	\label{watertank_objects_color_fls}
\end{figure*}

\section{Initial Results and Benchmarks}
We present a mix of new experimental results and analysis/comparison of previous results, mainly about baseline results for the different tasks and lessons for future research when using these datasets.

Table \ref{results_watertank} presents results for the Watertank dataset: object classification, patch matching, object detection with the Single Shot Multibox Detector (SSD) \cite{Liu_SSD}, and semantic segmentation with multiple segmentation and backbone networks. Object classification results on the Turntable dataset are presented in Table \ref{results_turntable}, and finally, self-supervised results on the Quarry dataset are shown in Table \ref{results_quarry}.

\begin{table*}[t]
	\begin{subtable}{0.25\linewidth}
		\begin{tabular}{ll}
			\toprule
			Model & Accuracy\\
			\midrule
			RBF SVM 	& 97.2 \%\\
			Linear SVM	& 97.5 \%\\
			Grad Boosting & 90.6\%\\
			Random Forest & 93.2\%\\
			ClassicCNN-BN & 99.2\%\\
			ClassicCNN-DO & 99.0\%\\
			FireNet & 99.7 \%\\
			\bottomrule
		\end{tabular}
		\caption{Object Classification Results}
	\end{subtable}
	\begin{subtable}{0.25\linewidth}
		\begin{tabular}{ll}
			\toprule
			Model & AUC\\
			\midrule
			SURF			& 67.9\%\\
			ORB 			& 68.2\%\\
			AKAZE			& 63.4\%\\
			SVM 			& 65.2\%\\
			Random Forest 	& 79.5\%\\
			2-Chan CNN 		& 91.0\%\\
			Siamese CNN 	& 85.5\%\\
			\bottomrule
		\end{tabular}
		\caption{Patch Matching Results}
	\end{subtable}
	\begin{subtable}{0.24\linewidth}
		\begin{tabular}{ll}
			\toprule
			SSD Backbone & mAP\\
			\midrule
			VGG16 @ $300 \times 300$& 91.69 \\
			ResNet20 & 89.85 \\
			MobileNet & 70.30\\
			DenseNet121 & 73.80\\
			SqueezeNet & 68.37 \\
			MiniXception & 71.62\\
			\bottomrule
		\end{tabular}
		\caption{Object Detection results.}
	\end{subtable}
	\begin{subtable}{0.24\linewidth}
		\begin{tabular}{lll}
			\toprule
			Seg Model & Backbone 		& mIoU\\
			\midrule
			UNet 		& RN34 & 74.8\% \\
			LinkNet 	& RN23 & 71.9\% \\
			DeepLabV3 	& RN50 & 66.8\%\\
			PSPNet 		& VGG16 & 71.3\%\\
			\bottomrule
		\end{tabular}
		\caption{Segmentation Results.}
	\end{subtable}
	\caption{Results on Watertank Dataset. Object detection results over different backbones using SSD, VGG16 uses $300 \times 300$ input size, while other backbones use $96 \times 96$ input size.}
	\label{results_watertank}
\end{table*}

Watertank results overall show that each successive task is more difficult than the previous one ($\text{OC} < \text{PM} < \text{OD} <\text{SS}$), shown by decreasing performance from object classification (easiest) to semantic segmentation (hardest), which makes sense as segmentation is a more complex task of predicting a dense mask output than object classification.

Our object classification results show that the Turntable dataset is more difficult compared to the Watertank dataset, with slightly lower accuracy on the Turntable dataset and requiring deeper models (like DenseNet) to achieve the best performance. Pre-training on the Turntable dataset and then performing transfer learning on the Watertank dataset via fine-tuning shows competitive performance but slightly worse than using the original Watertank training set.

We trained several self-supervised learning methods on the Quarry dataset, using the premade patches that we release in this paper, presented in Table \ref{results_quarry}. Feature learning from this dataset seems to be more difficult as performance is lower than supervised learning on the original Watertank training set, we believe this is because the self-supervised methods were developed for natural color images and not sonar ones, FLS-specific self-supervised methods should be developed in the future.

\textbf{Challenges}. The main challenge for most tasks is the trade-off between image size and performance. The larger the image, better performance, but this increases the required amounts of computation significantly. This is more relevant for object detection and segmentation tasks.

For object detection, the challenge is input image size, as SSD using VGG16 backbone at $300 \times 300$ input pixel resolution performs the best, with ResNet closely following, the problem being detection of small objects, which are common in marine debris.

For semantic segmentation, performance is lower than expected, and the biggest obstacle is the quality of labels, labeling was made from polygons, but this is not pixel-perfect. Sometimes sonar artifacts or background features are detected as objects, reducing overall mIoU.

For self-supervised learning, some methods like RotNet perform badly, probably because the task (predict a rotation angle) is not well defined for FLS images of the seafloor, there is too much ambiguitiy compared to natural color images. We motivate the need for sonar-specific self-supervised learning methods and pretext tasks.

\begin{table}[t]
		\begin{tabular}{lll}
			\toprule
							& \multicolumn{2}{c}{Accuracy (\%)}\\
			Model 			& Object Classification & TL to Watertank\\
			\midrule
			ResNet20 \cite{he2016deep}		& 93.6\% 	& 96.1\% \\
			MobileNet \cite{howard2017mobilenets} 		& 96.7\%	& 96.6\% \\
			DenseNet121 \cite{densenet}	&  99.4\%	& 98.0\%\\
			SqueezeNet \cite{iandola2016squeezenet}		& 99.1\%	& 97.5\% \\
			MiniXception \cite{arriaga2017real}	& 93.8\%	& 97.3\% \\
			Linear SVM		& 97.4\% 	& 92.9\% \\
			\bottomrule
		\end{tabular} 
	\caption{Results on the Turntable dataset for object classification and then transfer learning to the watertank dataset.}
	\label{results_turntable}
\end{table}

\begin{table}[t]
	\begin{tabular}{lll}
		\toprule
		Model 			& SSL Task Performance & WT TL Acc (\%)\\
		\midrule
		SVM Baseline 	& NA 		& 95.7\%\\
		RotNet 			& 95\% Acc 	& 12.7\%\\
		Denoising Autoencoder & 0.12 MAE	& 85.2\%\\
		Jigsaw Puzzle 	& 97\% Acc & 97.0\%\\
		\bottomrule
	\end{tabular} 
	\caption{Self-Supervised learning on the Quarry dataset and Transfer Learning to the Watertank Dataset results.}
	\label{results_quarry}
\end{table}

\section{Conclusions and Future Work}
In this paper we have presented our marine debris FLS datasets containing multiple sonar image capture settings and multiple computer vision tasks, and we provide a unified description of all datasets and introduce new sub-datasets with new data and object classes. We provide initial benchmark results, showing the difficulty of the tasks and lessons for future research.

We expect that our datasets, in particular the Quarry datasets for unsupervised learning, will be used by the community to 
build more advanced machine learning and computer vision models and move research on sonar image forward.

We have fully publicly released all datasets, in particular the Quarry dataset was not previously publicly available, while we have full releases for all other scenarios. The dataset is available in Github and Zenodo at \url{https://doi.org/10.5281/zenodo.15101686}.

\textbf{Limitations}. While the datasets presented in this paper are the largest publicly available FLS datasets, they still lag in size compared with computer vision datasets (in the range of millions of images), and our datasets range in thousands of images, limiting the usefulness for machine learning models. We use a limited set of objects to generate data, and more variability is left for future work. The Quarry dataset enables unsupervised learning and many models and unsupervised learning methods could be tried in the future, we only tested a handful of self-supervised models. Segmentation labels are built from rasterized polygons and might not be precise enough for some applications.

\textbf{Other Tasks}. Our datasets can be used for many more tasks that we did not explore, some examples are: combining bounding boxes and masks for instance segmentation, patch matching in the turntable dataset, and all datasets could be used for evaluation of out of distribution detection.

\section*{Acknowledgments}
The authors would like to thank Leonard McLean for his help in capturing data used in this paper, this dataset would not be possible without his help.

\FloatBarrier

\bibliographystyle{ieeetr} %
\bibliography{biblio.bib}

\appendices
\section{Broader Impact Statement}
Datasets are used as a proxy for reality when used to train machine learning and computer vision models. The datasets presented in this paper have academic value to build and evaluate new models, but they likely do not fully generalize to real-world underwater environments. Every dataset is biased and does not fully represent reality.

The Watertank and Turntable datasets contain object classes for marine debris objects, and it is only usable in that context, and while we expect that many kind of models (like feature learning, unsupervised and self-supervised learning) would generalize outside the context of marine debris, there are no performance guarantees and extensive verification should be done in each case.

\section{Additional Dataset Information and Statistics}

This section provides additional information and statistics. The three scenarios are meant to provide data for academic research in underwater perception, in particular for machine learning and computer vision model training and development. It is not expected that models trained on these datasets will generalize to reality.

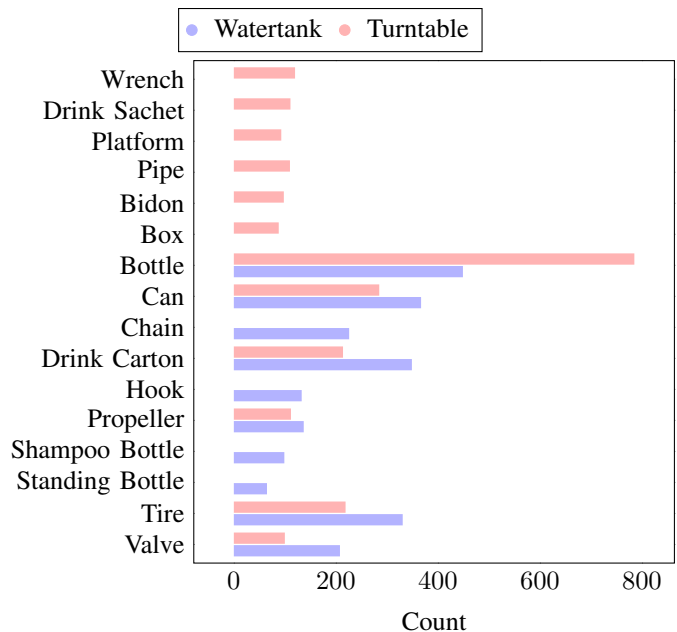
\begin{figure}[t]
	\centering
	\begin{tikzpicture} 
		\begin{customlegend}[legend columns = 2,legend style = {column sep=1ex}, legend cell align = left,
			legend entries={Watertank, Turntable}]
			\addlegendimage{mark=none,blue!30, only marks}
			\addlegendimage{mark=none,red!30, only marks}
		\end{customlegend}
	\end{tikzpicture}
	\vspace*{0.3em}
	
	\begin{tikzpicture} 
		\begin{axis}[xbar=0.5pt, enlarge y limits=0.04, xlabel={Count}, symbolic y coords={ 
				Valve, Tire, Standing Bottle, Shampoo Bottle, Propeller, Hook, Drink Carton, Chain, Can, Bottle, Box, Bidon, Pipe, Platform, Drink Sachet, Wrench}, ytick=data, nodes near coords align={horizontal}, height = 0.35 \textheight, width=0.9\linewidth, tick align=inside, tickwidth=0pt, bar width=0.15cm]
			\addplot[draw=none, fill=blue!30]coordinates {
				(449,Bottle) (367,Can) (226,Chain) (349,Drink Carton) (0,Drink Sachet) (133,Hook)
				(137,Propeller) (99,Shampoo Bottle) (65,Standing Bottle) (331,Tire)
				(208,Valve) (0,Bidon) (0,Box) (0,Pipe) (0,Platform)  (0,Wrench)
			};
			\addplot[draw=none, fill=red!30] coordinates {
				(785,Bottle) (98,Bidon) (88,Box) (285,Can) (214,Drink Carton) (111,Drink Sachet) (110,Pipe) (93,Platform) (112,Propeller)  (219,Tire) (100,Valve) (120,Wrench)
			};
		\end{axis}
	\end{tikzpicture}
	\caption{Distribution of Class Samples for the Turntable and Watertank Datasets}
	\label{dataset_histogram}
\end{figure}

Figure \ref{dataset_histogram} shows the class distribution for Watertank and Turntable scenarios, while both datasets have imbalances, the Turntable dataset is more balanced except for the bottle class. When training baseline models, imbalance was not a big factor during training and no corrections for imbalance were required.

Table \ref{class_descriptions} provides short description for the object classes used to build the Watertank and Turntable datasets.

\begin{figure*}
	\begin{subfigure}{0.49\linewidth}
		\includegraphics[width=0.24\linewidth]{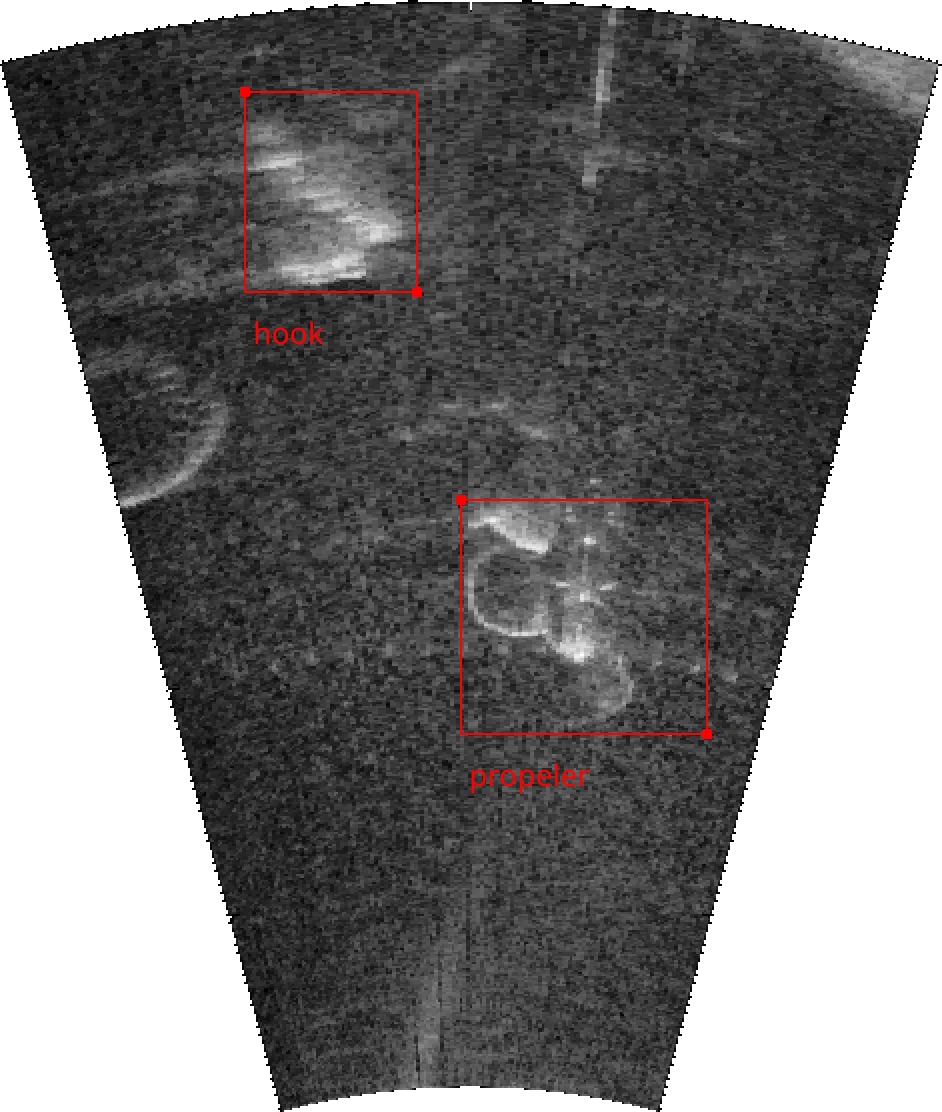}
		\includegraphics[width=0.24\linewidth]{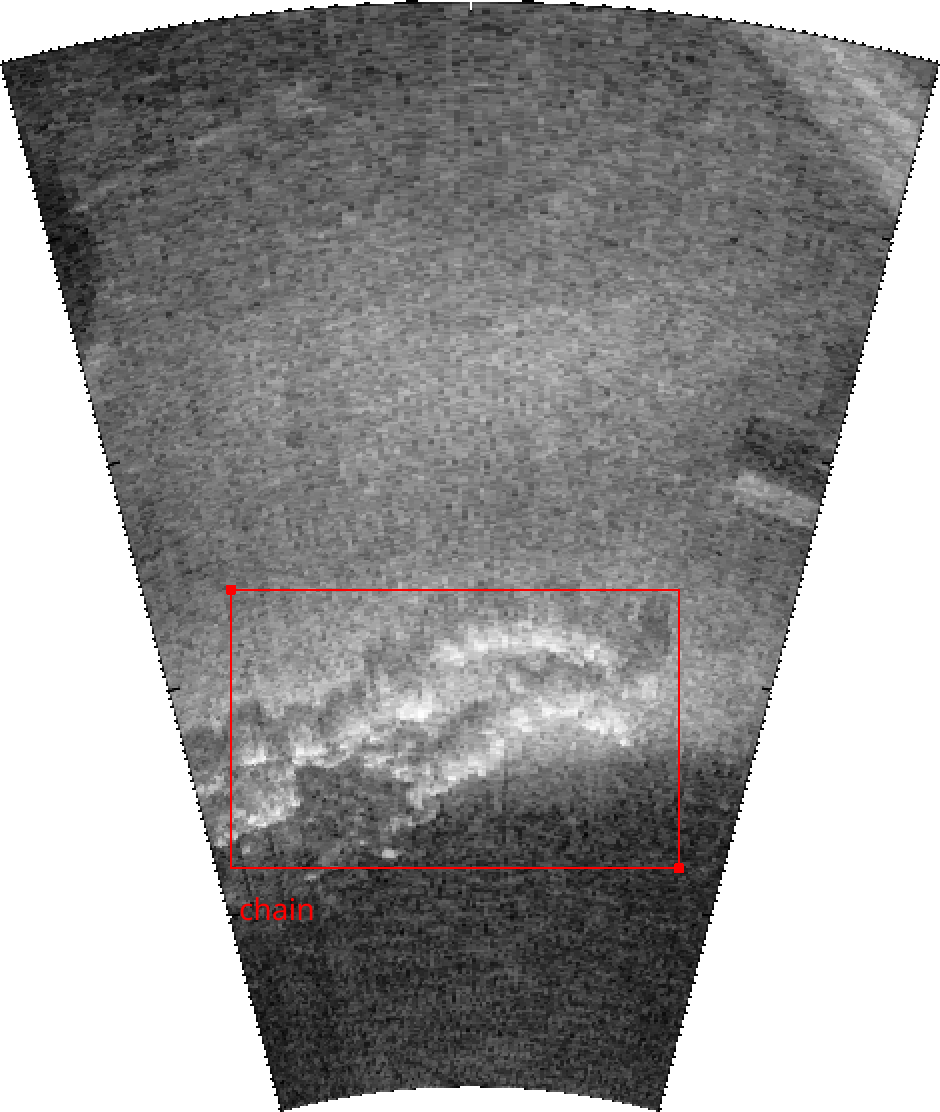}
		\includegraphics[width=0.24\linewidth]{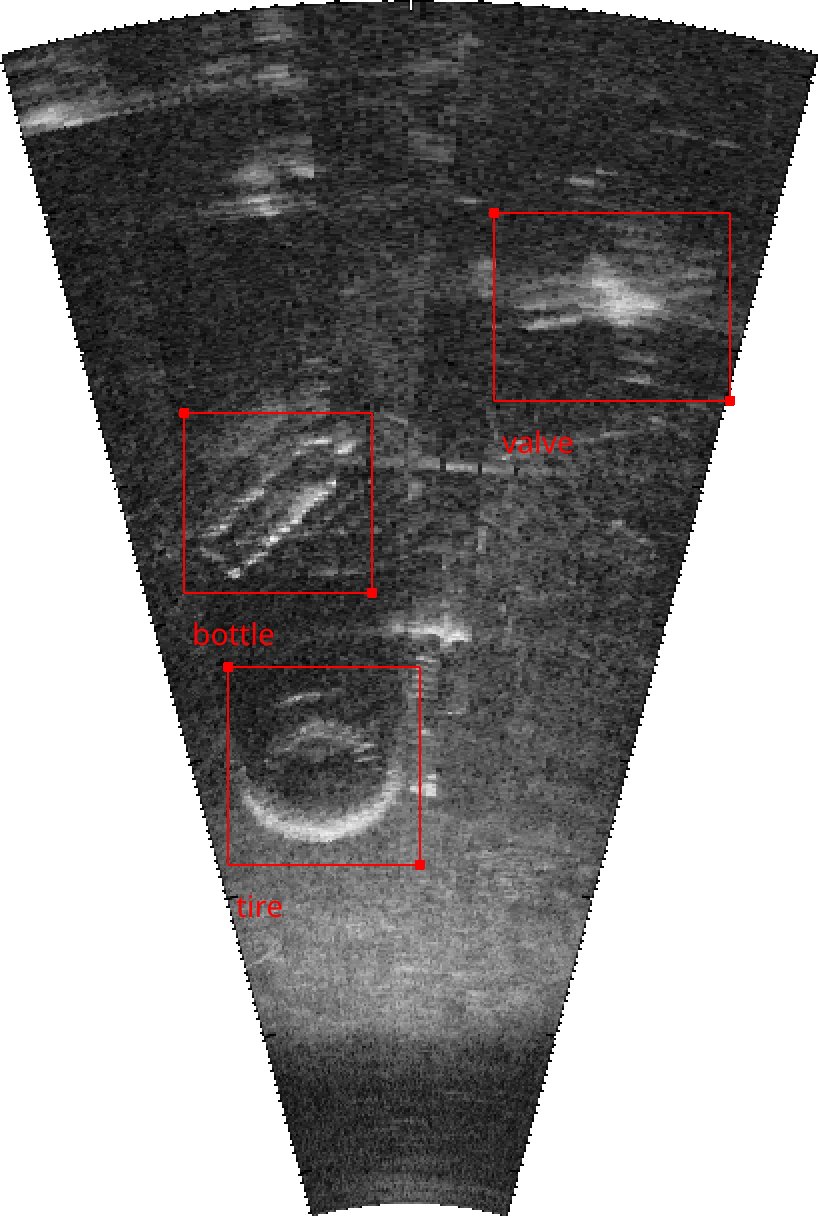}		
		\includegraphics[width=0.24\linewidth]{images/watertank/sonar-scene-bottles.png}
		\caption{Bounding Box}
	\end{subfigure}
	\begin{subfigure}{0.49\linewidth}		
		\includegraphics[width=0.24\linewidth]{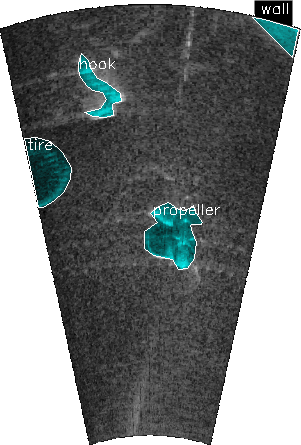}
		\includegraphics[width=0.24\linewidth]{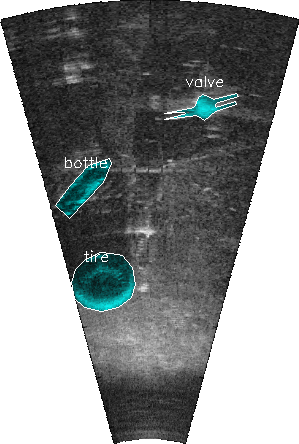}
		\includegraphics[width=0.24\linewidth]{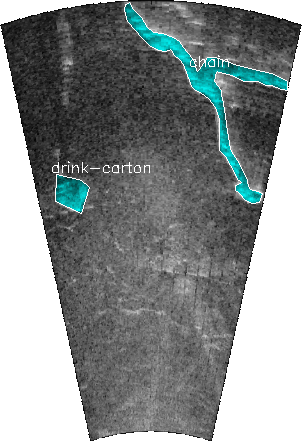}
		\includegraphics[width=0.24\linewidth]{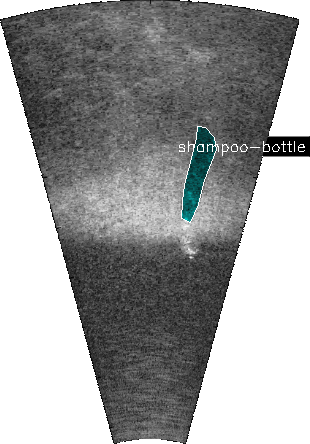}
		\caption{Semantic Segmentation}
	\end{subfigure}
	\caption{Examples of bounding box and semantic segmentation labels in the watertank dataset.}
	\label{watertank_label_examples}
\end{figure*}

\begin{figure*}[!h]
	\centering
	\begin{tabular}{rlp{5cm}p{3cm}p{3cm}}
		\toprule
		& Object Sample & \multicolumn{2}{c}{FLS Sample}\\
		Object Class 	& Image & Watertank View & Turntable Sideways & Turntable Standing\\
		\midrule
		Bidon 		& \sampimgtab{images/turntable/bidon.jpg} & NA &
		\sampimgtab{images/turntable/classes/plastic-bidon-sideways.png} \\
		Bottle 			& \sampimgtab{images/watertank/glass-bottle.jpg} & \sampimgtab{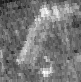} \sampimgtab{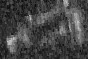} \sampimgtab{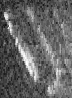} &
		\sampimgtab{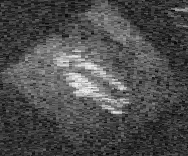}  & \sampimgtab{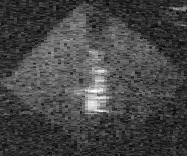}\\
		Can 			& \sampimgtab{images/watertank/can.jpg} & \sampimgtab{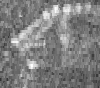} \sampimgtab{images/watertank/classes/can-1.png} \sampimgtab{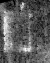} &
		\sampimgtab{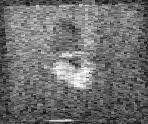} & \sampimgtab{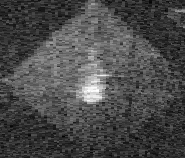} \\
		Chain 			& \sampimgtab{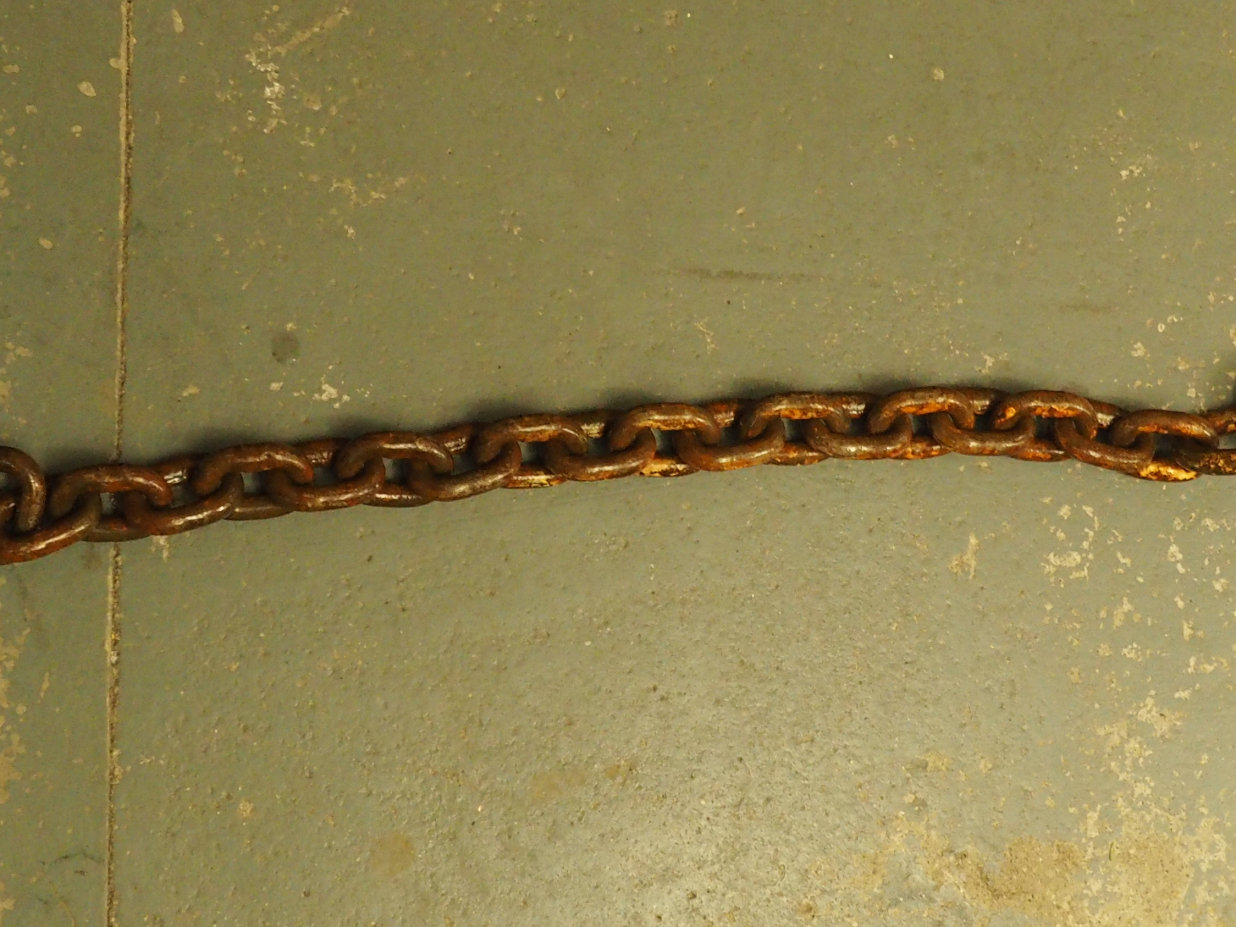} & \sampimgtab{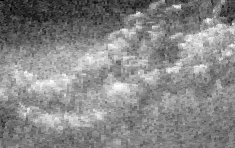} \sampimgtab{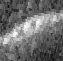} &  &  \\
		\shortstack{Drink\\Carton}	& \sampimgtab{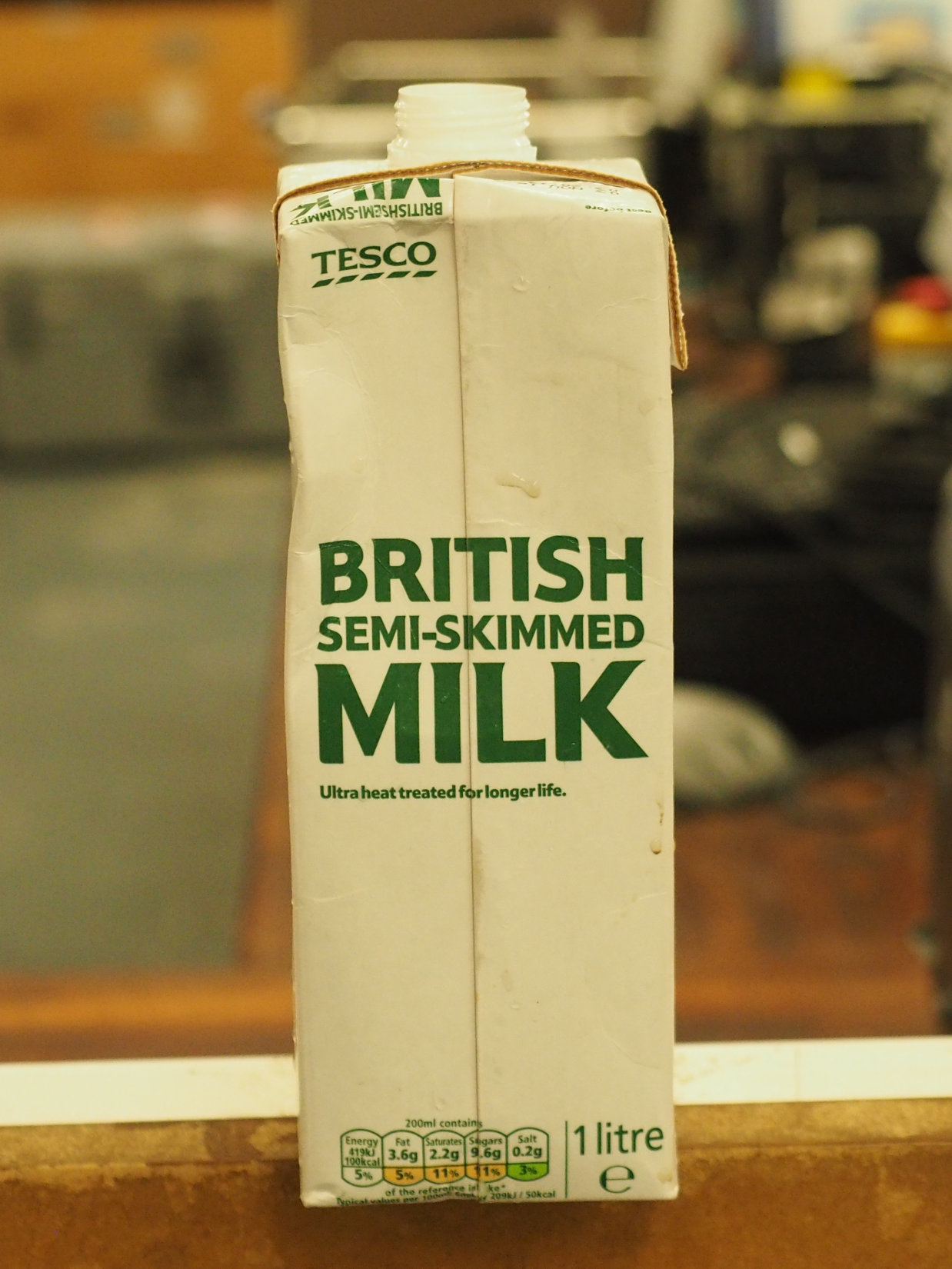} & \sampimgtab{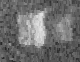} \sampimgtab{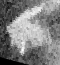} \sampimgtab{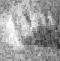} &
		\sampimgtab{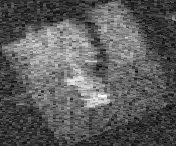} & \sampimgtab{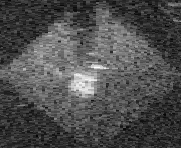} \\
		\shortstack{Drink\\Sachet} & \sampimgtab{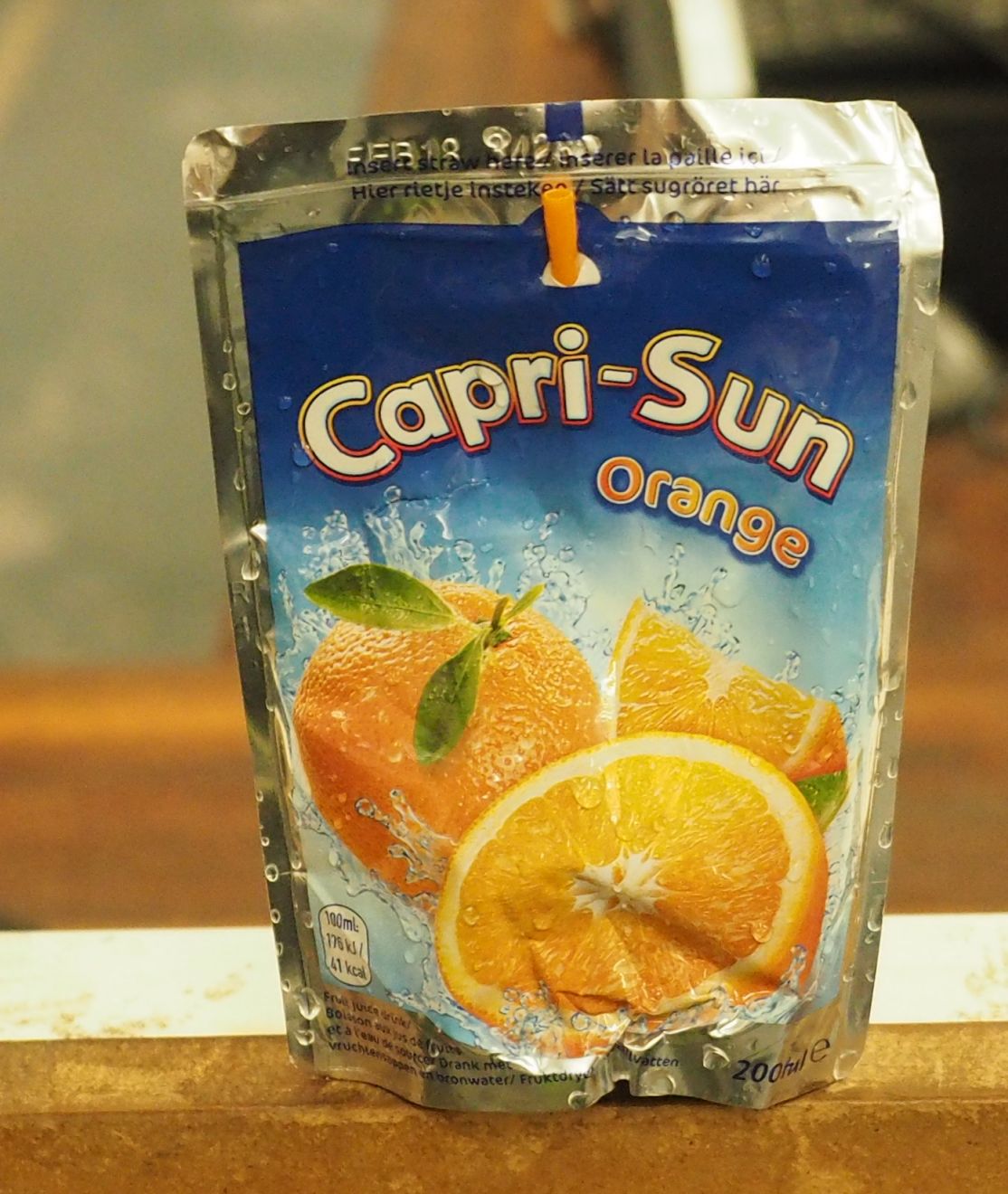} &   &  &
		\sampimgtab{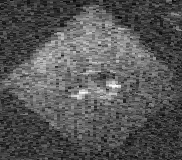} \\
		Hook 			& \sampimgtab{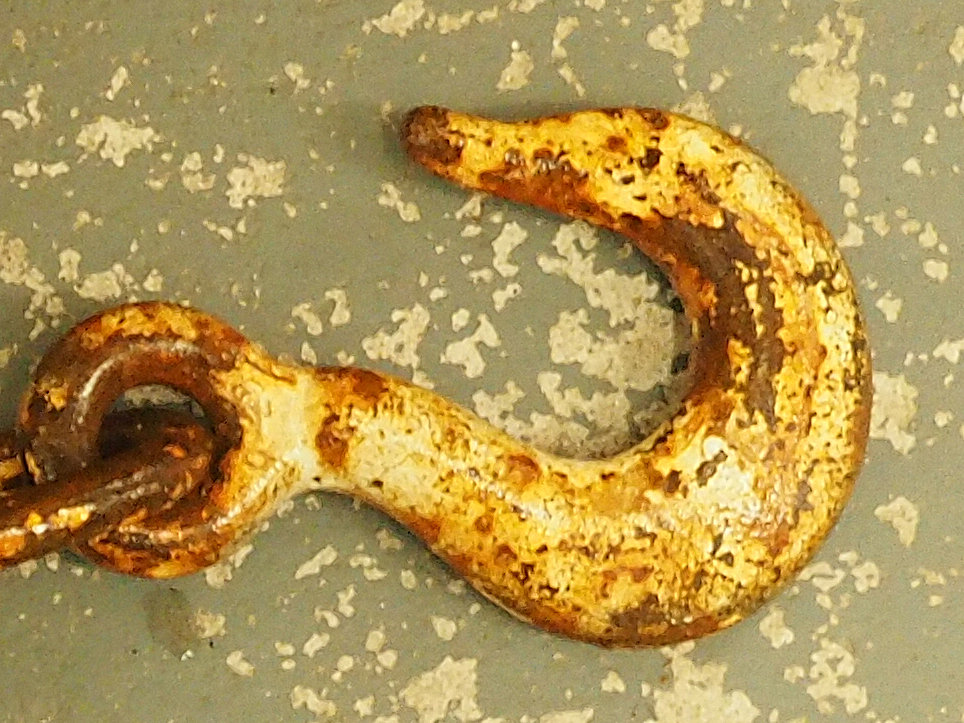} & \sampimgtab{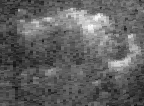} \sampimgtab{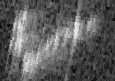} & \\
		Jar			& \sampimgtab{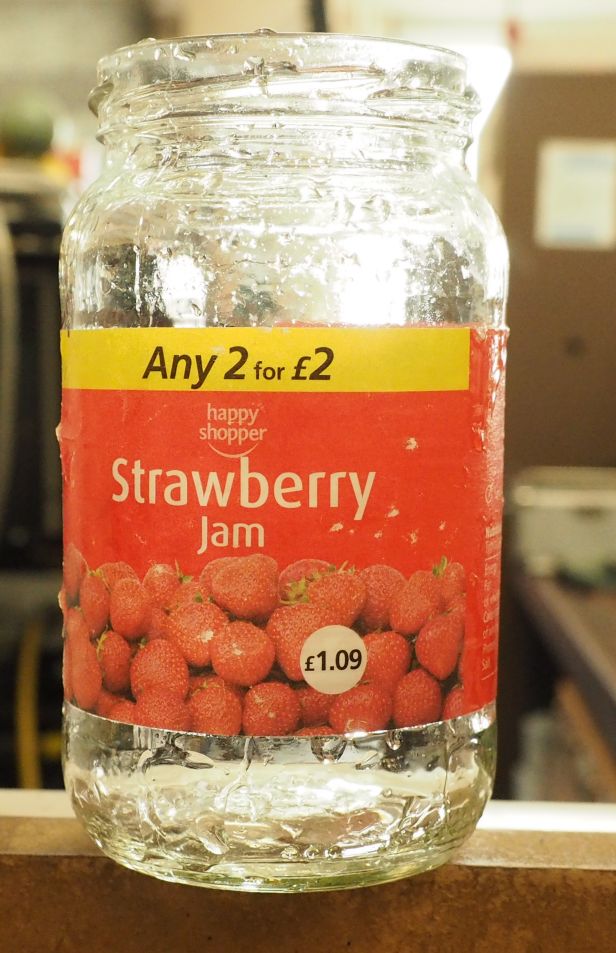}&   & \sampimgtab{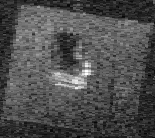} &
		\sampimgtab{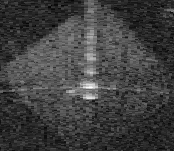} \\        
		\shortstack{Metal\\Box}			& \sampimgtab{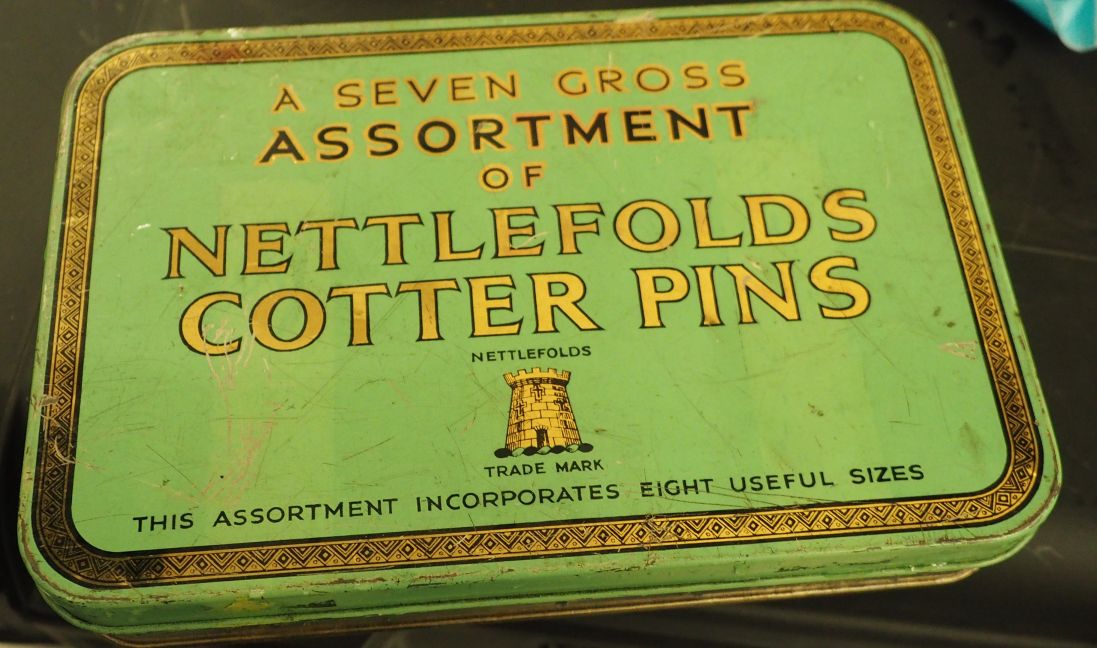}&   &  &
		\sampimgtab{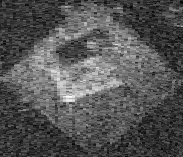} \\
		Pipe		& \sampimgtab{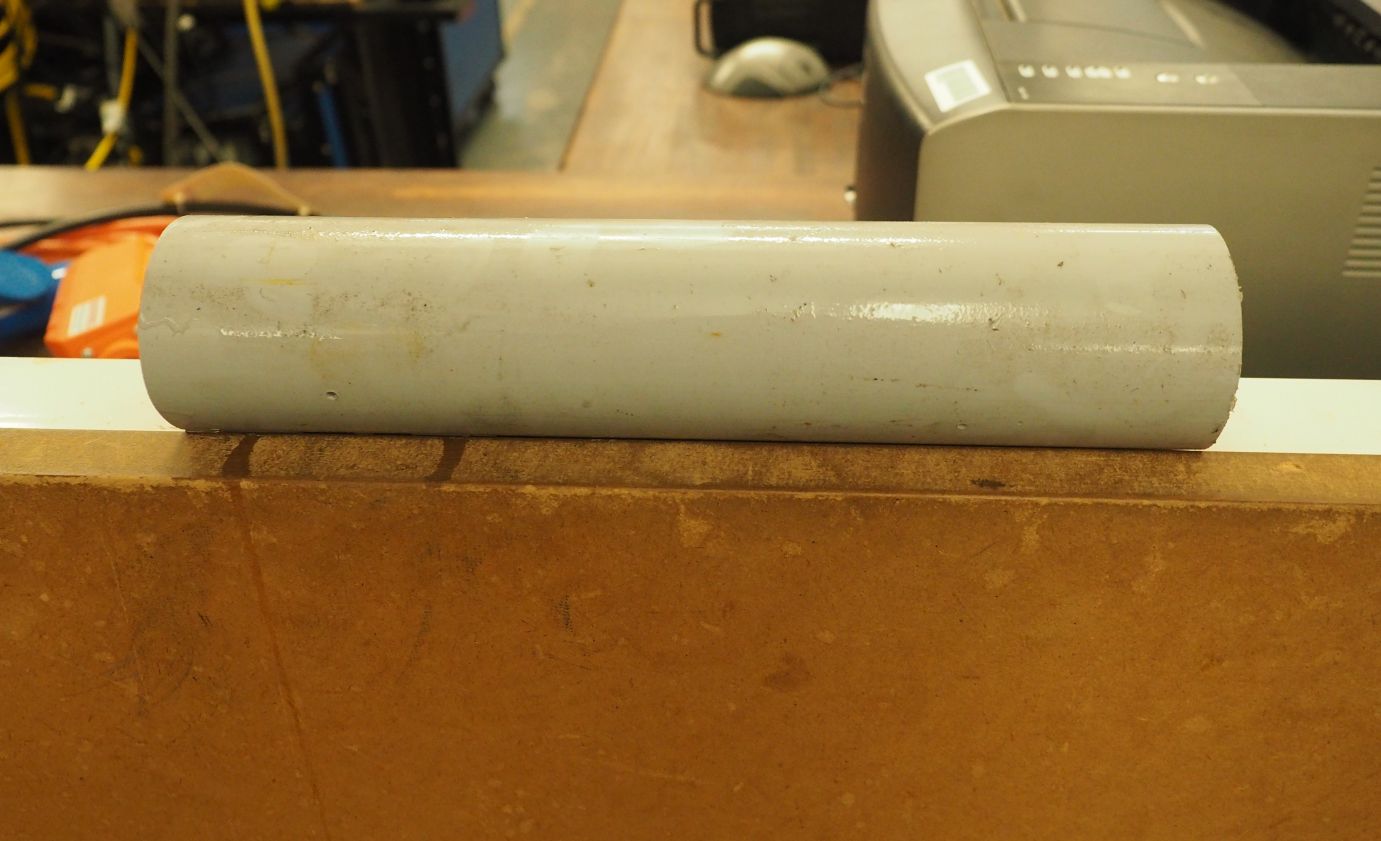} &  &
		\sampimgtab{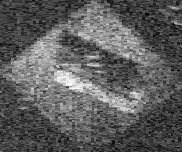} \\                
		Platform		& \sampimgtab{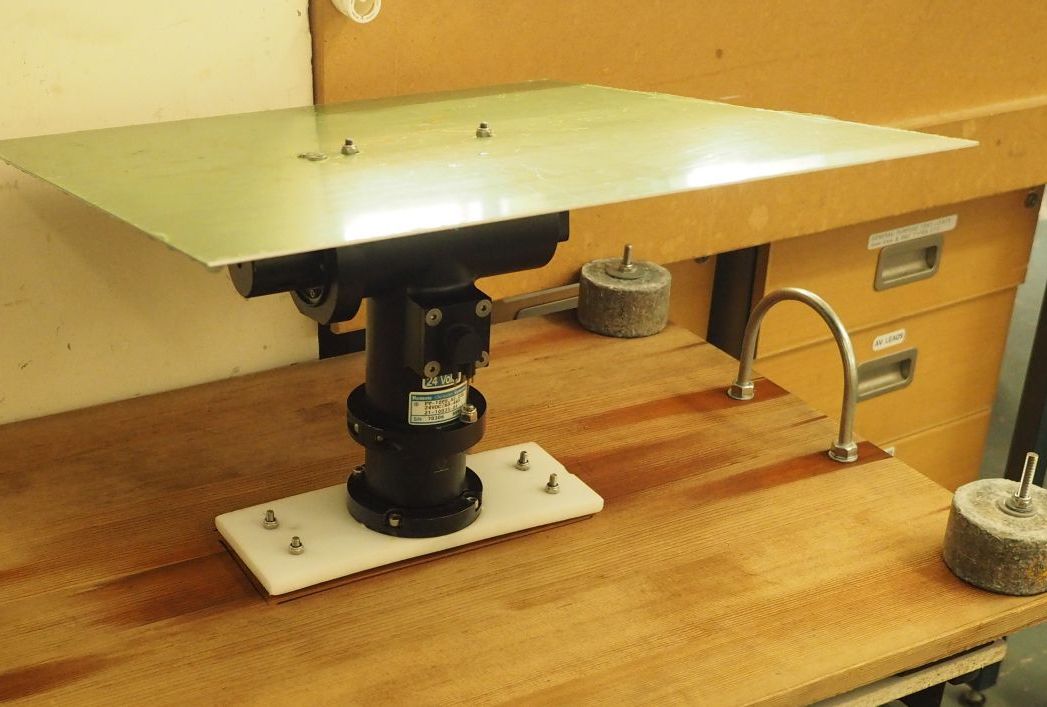} &  &        \sampimgtab{images/turntable/classes/rotating-platform-standing.png} \\
		\shortstack{Plastic\\Bottle}	& \sampimgtab{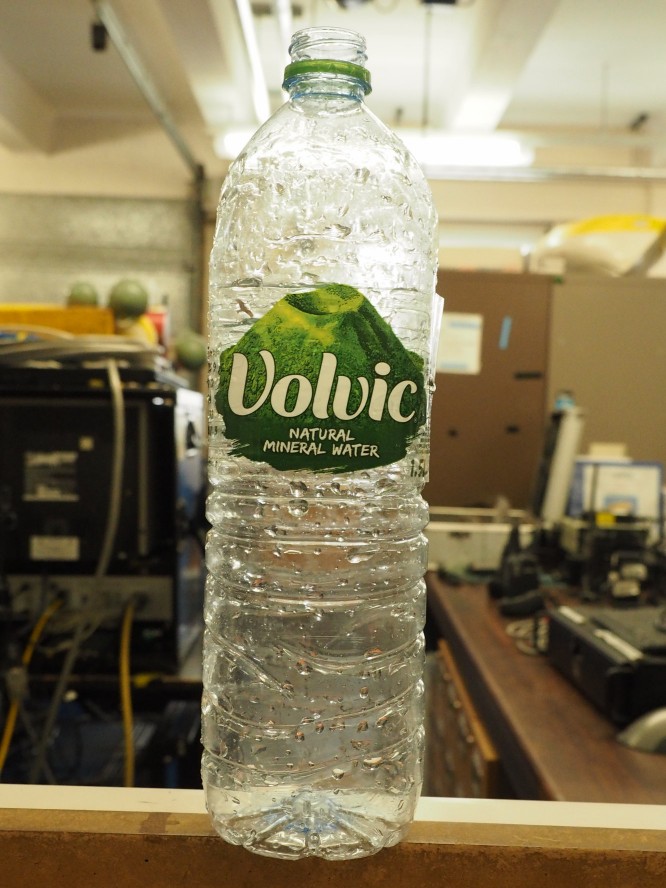} &   & \sampimgtab{images/turntable/classes/plastic-bottle-sideways.png} &
		\sampimgtab{images/turntable/classes/plastic-bottle-standing.png} \\
		Propeller		& \sampimgtab{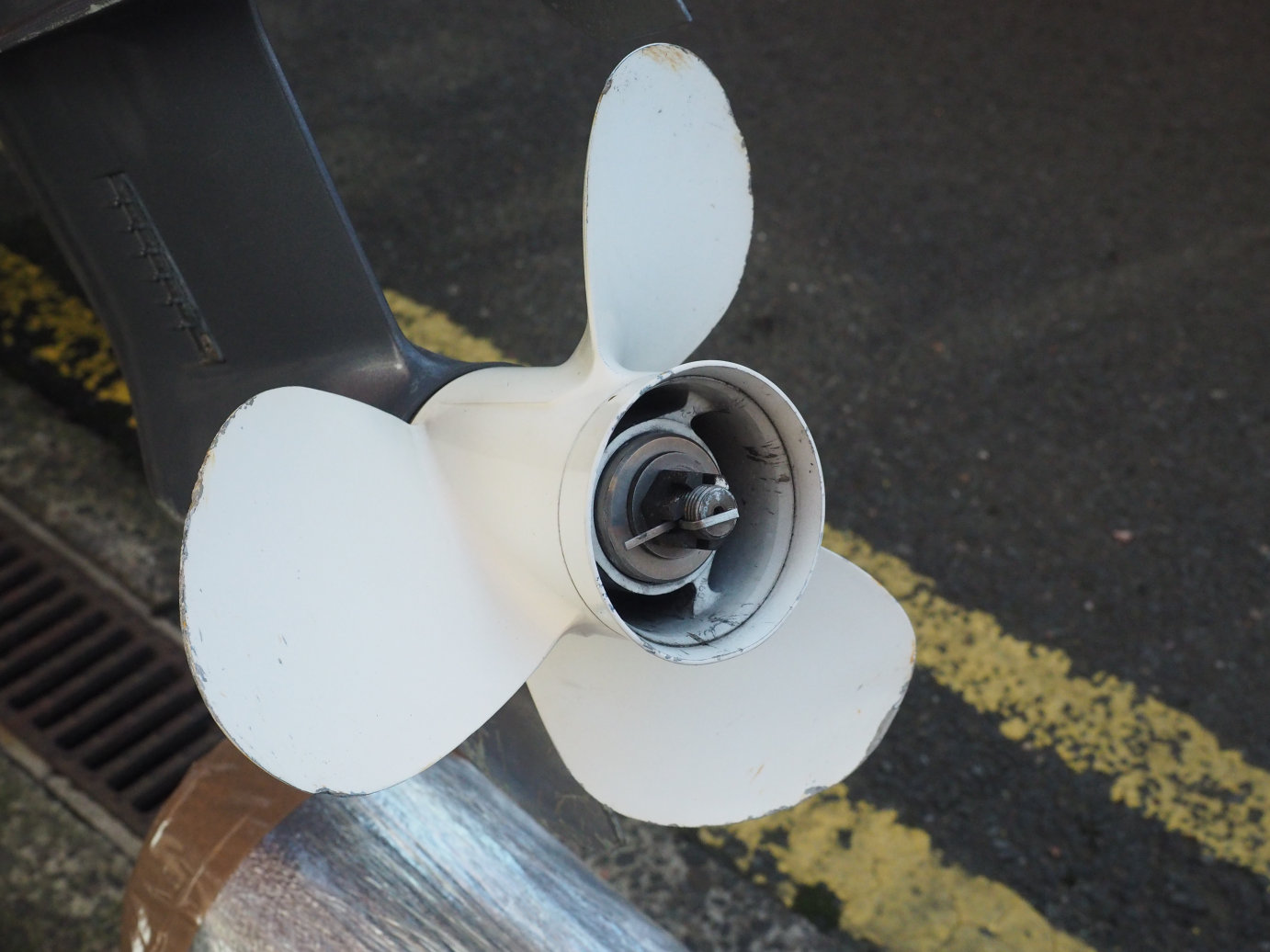} & \sampimgtab{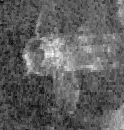} \sampimgtab{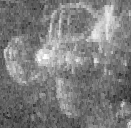} \sampimgtab{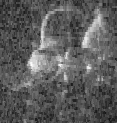} &
		\sampimgtab{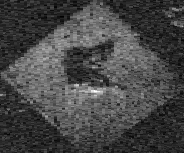} \\
		\shortstack{Shampo\\Bottle}	& \sampimgtab{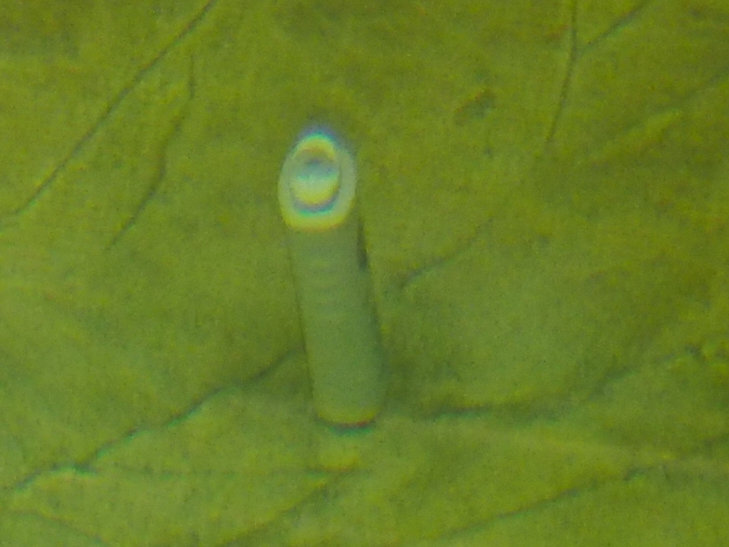} & \sampimgtab{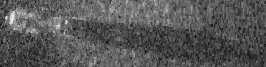} &  \\
		\shortstack{Standing\\Bottle}	& \sampimgtab{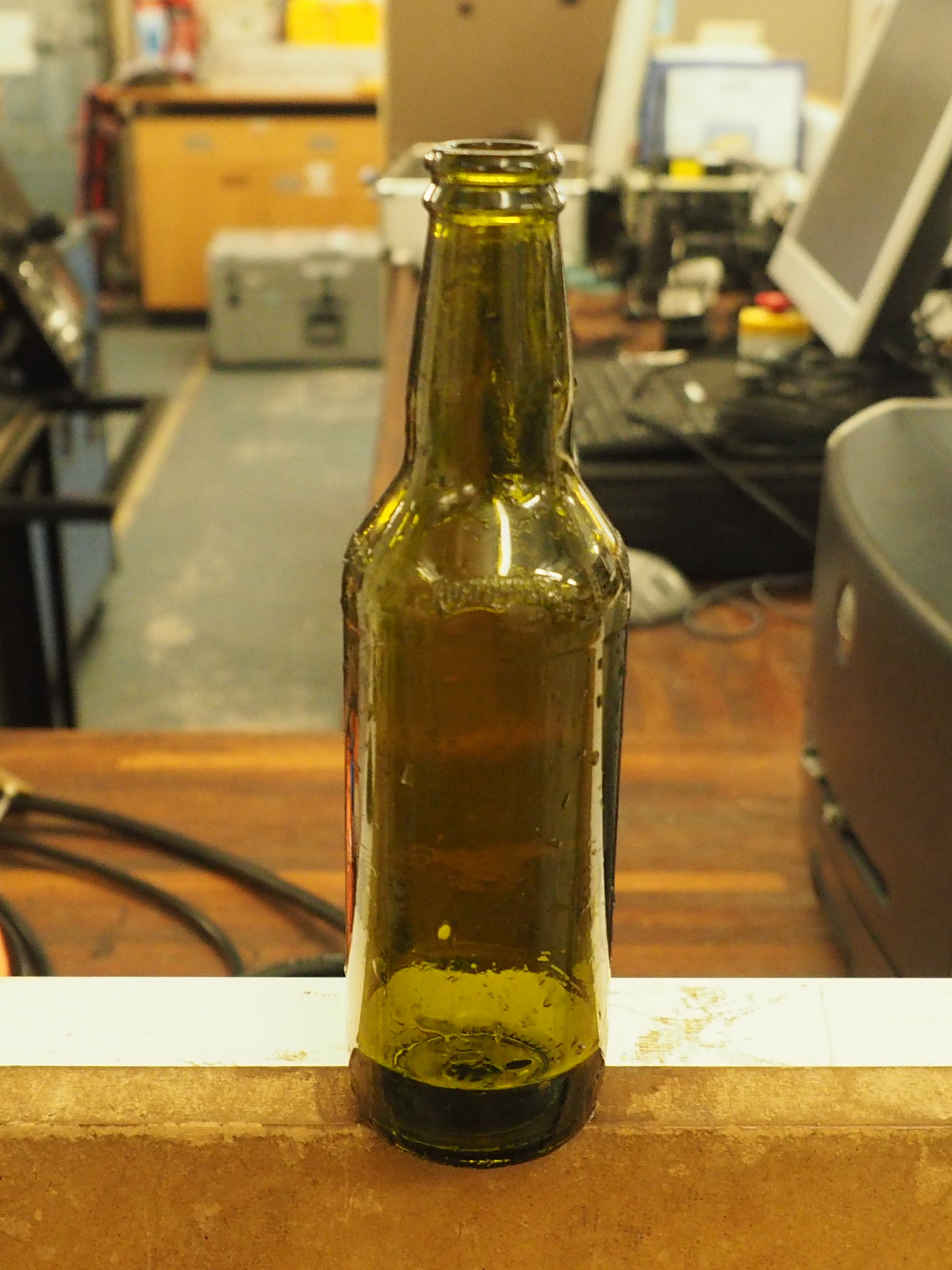} & \sampimgtab{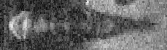} & \\
		Tire			& \sampimgtab{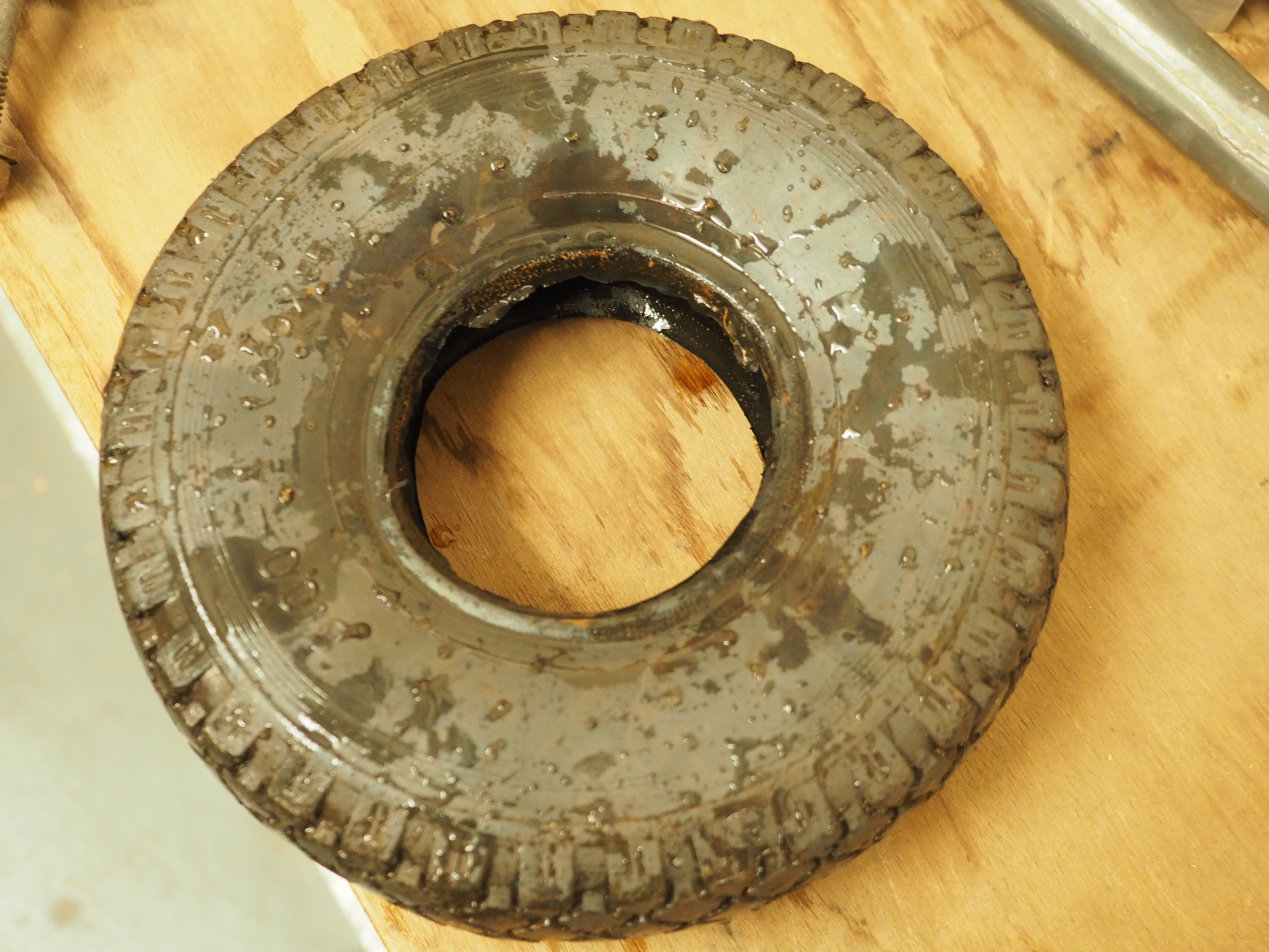} & \sampimgtab{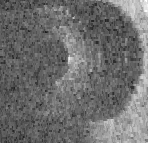} \sampimgtab{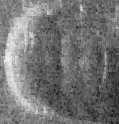} \sampimgtab{images/watertank/classes/tire-2.png} &
		\sampimgtab{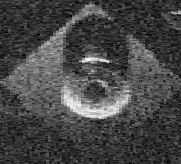} \\
		Valve			& \sampimgtab{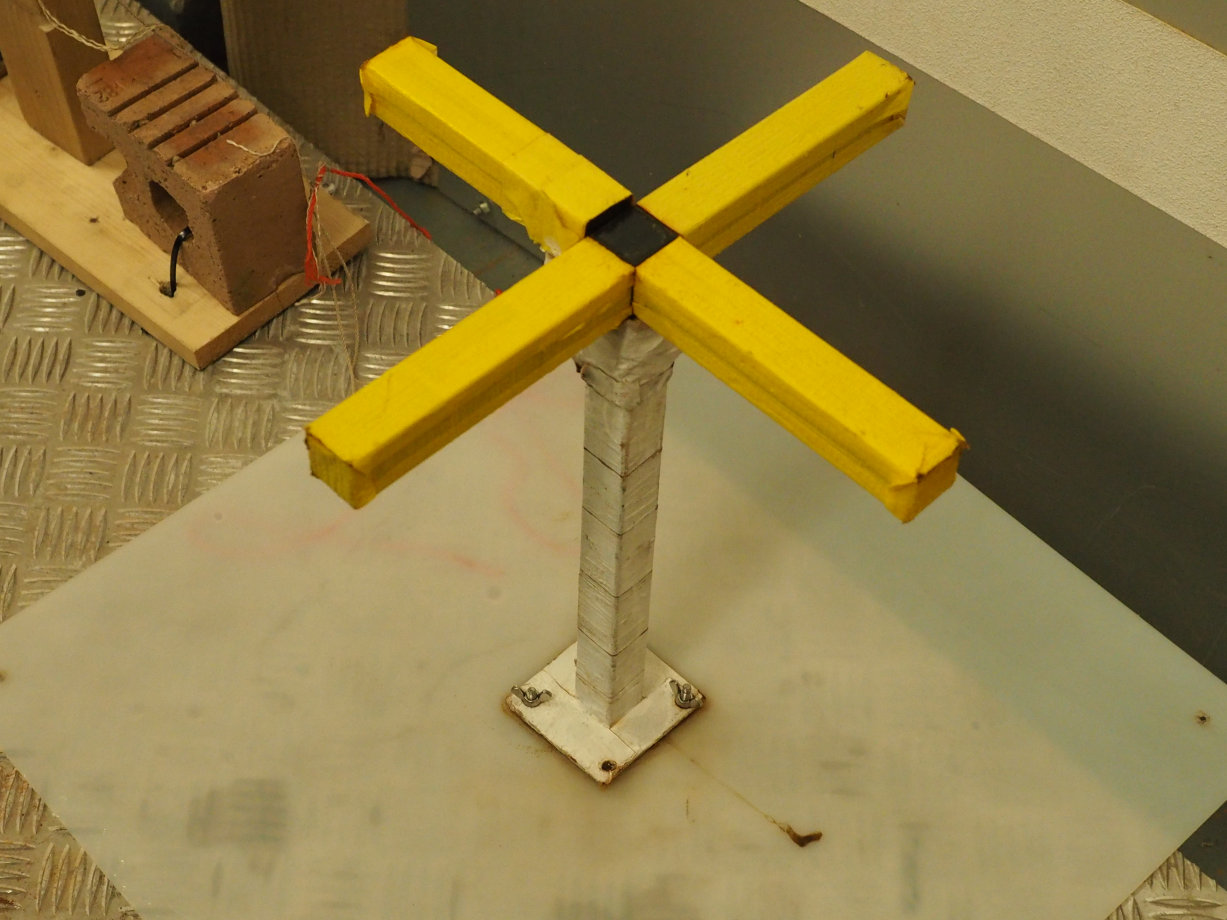} & \sampimgtab{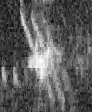} \sampimgtab{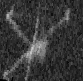} \sampimgtab{images/watertank/classes/valve-2.png} &
		\sampimgtab{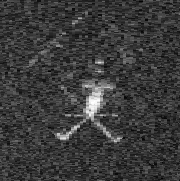} \\                        
		Wrench & \sampimgtab{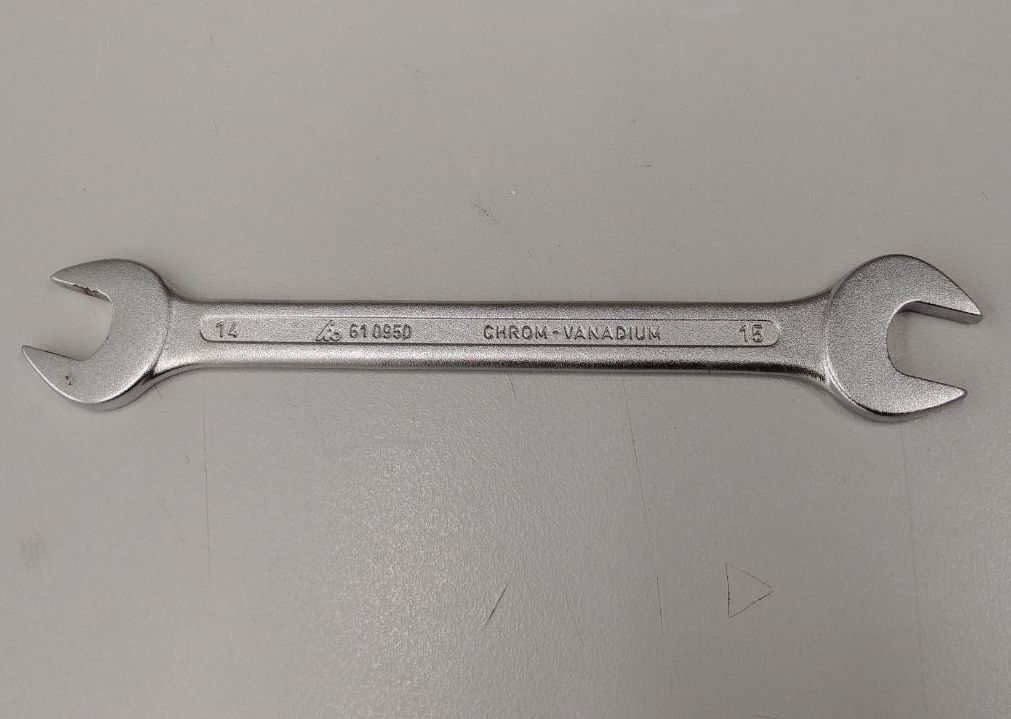} &   &  &
		\sampimgtab{images/turntable/classes/wrench-standing.png} \\        
		\bottomrule
	\end{tabular}
	\caption{Comparison of objects in the Watertank and Turntable datasets, including a photo of each real-world objects in air. Note that not every object is available in each dataset. Empty images indicate that the object is not present in that combination.}
	\label{big_fls_color_objects}
\end{figure*}

Figure \ref{watertank_label_examples} presents sample bounding box and semantic segmentation annotations in full size FLS images, and Figure \ref{big_fls_color_objects} presents a large summary of object classes and Watertank/Turntable FLS views.

\begin{figure}
	\begin{subfigure}{\linewidth}
		\includegraphics[width=0.19\linewidth]{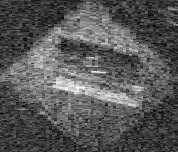}
		\includegraphics[width=0.19\linewidth]{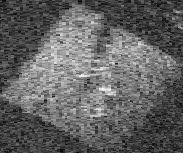}
		\includegraphics[width=0.19\linewidth]{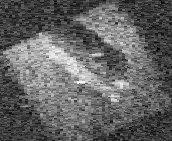}
		\includegraphics[width=0.19\linewidth]{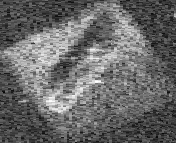}
		\includegraphics[width=0.19\linewidth]{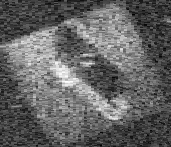}
		\caption{Pipe}
	\end{subfigure}
	\begin{subfigure}{\linewidth}
		\includegraphics[width=0.19\linewidth]{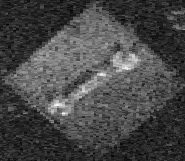}
		\includegraphics[width=0.19\linewidth]{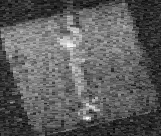}
		\includegraphics[width=0.19\linewidth]{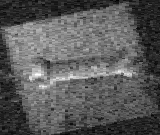}
		\includegraphics[width=0.19\linewidth]{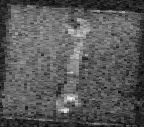}
		\includegraphics[width=0.19\linewidth]{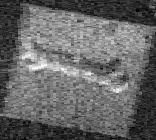}
		\caption{Wrench}
	\end{subfigure}
	\begin{subfigure}{\linewidth}
		\includegraphics[width=0.19\linewidth]{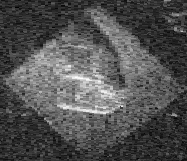}
		\includegraphics[width=0.19\linewidth]{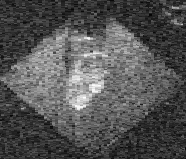}
		\includegraphics[width=0.19\linewidth]{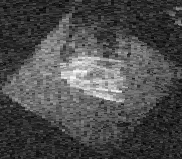}
		\includegraphics[width=0.19\linewidth]{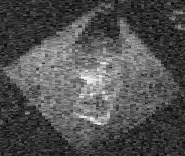}
		\includegraphics[width=0.19\linewidth]{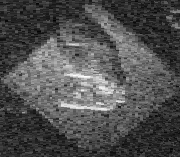}
		\caption{Bidon}
	\end{subfigure}
	\begin{subfigure}{\linewidth}
		\includegraphics[width=0.19\linewidth]{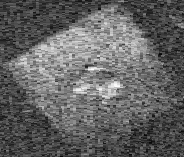}
		\includegraphics[width=0.19\linewidth]{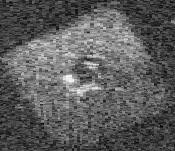}
		\includegraphics[width=0.19\linewidth]{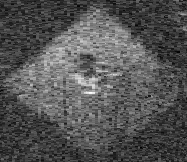}
		\includegraphics[width=0.19\linewidth]{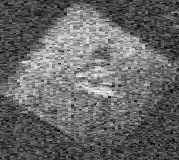}
		\includegraphics[width=0.19\linewidth]{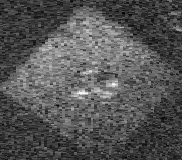}
		\caption{Drink Sachet}
	\end{subfigure}
	\caption{Additional turntable rotating examples.}
	\label{turntable_add_examples}
\end{figure}

Figure \ref{turntable_add_examples} presents additional FLS examples of the rotating turntable with multiple objects.

\begin{table*}[t]
	\centering
	\begin{subtable}{0.49\textwidth}
		\begin{tabular}{llp{4.5cm}}
			\toprule
			ID 			& Name 				& Description\\ 
			\midrule
			0			& Bottle			& Plastic and Glass bottles, lying horizontally\\
			1			& Can				& Several metal cans originally containing food \\
			2			& Chain				& A one meter chain with small chain links\\
			3			& Drink Carton		& Several milk/juice drink cartons lying horizontally\\
			4			& Hook				& A small metal hook \\
			5			& Propeller			& A small ship propeller made out of metal \\
			6			& Shampoo Bottle	& A standing plastic shampoo bottle  \\
			7			& Standing Bottle	& A standing beer bottle made out of glass \\
			8			& Tire				& A small rubber tire lying horizontally \\
			9			& Valve				& A mock-up metal valve originally designed for the euRathlon 2015 competition\\
			10			& Background		& Anything that is not an object, usually the bottom of our water tank\\
			\bottomrule
		\end{tabular}
		\caption{Watertank Scenario}
	\end{subtable}
	\begin{subtable}{0.49\textwidth} 
		\begin{tabular}{lp{3cm}ll}
								&													& \rotatebox{45}{Standing} & \rotatebox{45}{Sideways}\\
			\toprule
			Class/Object Name	& Description										& \multicolumn{2}{c}{Object Poses}\\
			\midrule
			Bottle				& Bottles made out of plastic, metal, glass, etc	& \cmark & \cmark\\
			Can					& A metallic can									& \cmark & \cmark\\
			Drink Carton		& A milk carton										& \cmark & \cmark\\
			Drink Sachet		& A "Capri Sun" plastic drink sachet				& \xmark & \cmark\\
			Jar					& A glass jam jar									& \cmark & \cmark\\
			Box					& A metallic box									& \cmark & \xmark\\
			Tire				& One large and one small rubber tire				& \xmark & \cmark\\
			Bidon				& A plastic milk bidon								& \xmark & \cmark\\
			Pipe				& A short plastic pipe								& \xmark & \cmark\\
			Propeller			& A plastic boat propeller							& \cmark & \xmark\\
			Platform			& The rotating platform, without an object			& \cmark & \xmark\\
			Valve				& A mockup of an underwater valve					& \cmark & \xmark\\
			Wrench				& A metal wrench									& \cmark & \xmark\\
			\bottomrule
		\end{tabular}
		\caption{Turntable Scenario}
	\end{subtable}
	\caption{Description of object classes available in the Watertank and Turntable scenarios, including pose information for Turntable objects.}
	\label{class_descriptions}
\end{table*}

\section{Additional Failed FLS Images from Quarry Scenario}

\begin{figure}
	\centering
	\includegraphics[width=0.18\linewidth]{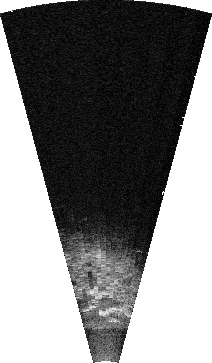}
	\includegraphics[width=0.18\linewidth]{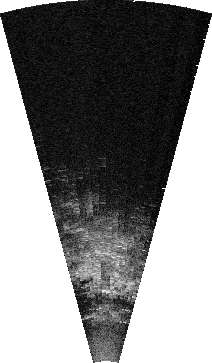}
	\includegraphics[width=0.18\linewidth]{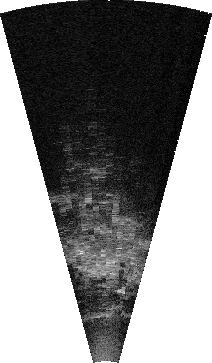}
	\includegraphics[width=0.18\linewidth]{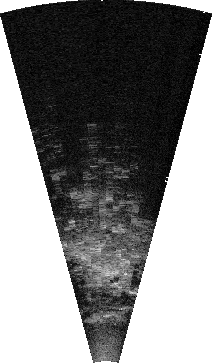}
	\caption{Examples of failed sonar images due to lateral movement, showing beam interference.}
	\label{failed_fls}
\end{figure}

When capturing data in the Quarry scenario, we used Sonobot with the FLS attached in the bottom. As the Sonobot is a surface vehicle, due to wind and waves in the water, the Sonobot would shake laterally, which would produce FLS images that have beam interference, which produced FLS images that are not useful. These images are presented in Figure \ref{failed_fls}.

\end{document}